\documentclass[11pt]{article}
\usepackage[preprint]{acl}

\usepackage{times}
\usepackage{latexsym}
\usepackage[T1]{fontenc}
\usepackage[utf8]{inputenc}
\usepackage{microtype}
\usepackage{inconsolata}

\usepackage{graphicx}
\usepackage{amsmath}
\usepackage{amsfonts}
\usepackage{amsthm}
\usepackage{booktabs}
\usepackage{multirow}
\usepackage{xcolor}
\usepackage{pifont}
\usepackage{colortbl}
\usepackage[most]{tcolorbox}
\tcbuselibrary{skins,breakable,listings}
\usepackage{enumitem}
\usepackage{subcaption}
\usepackage{listings}

\newtcolorbox{promptbox}{
  colback=gray!3,
  colframe=gray!60,
  boxrule=0.6pt,
  arc=2mm,
  left=6pt,
  right=6pt,
  top=6pt,
  bottom=6pt,
  breakable
}

\definecolor{acadblue}{RGB}{0,51,102}
\definecolor{acadgray}{RGB}{245,245,245}
\definecolor{constraintred}{RGB}{217,83,79}
\definecolor{inputgray}{RGB}{245,245,245}
\definecolor{modelblue}{RGB}{235,245,255}
\definecolor{labelgreen}{RGB}{235,255,235}

\definecolor{roiGreen}{HTML}{187765}
\definecolor{roiGray}{HTML}{8E959C}
\definecolor{roiTint}{HTML}{EDF4F8}
\newcommand{\roiyes}{\textcolor{roiGreen}{\ding{51}}}
\newcommand{\roino}{\textcolor{roiGray}{\ding{55}}}

\title{ROI-Reasoning: Rational Optimization for Inference via Pre-Computation Meta-Cognition}

\author{
    \textbf{Muyang Zhao}\textsuperscript{1}\thanks{Equal contribution.},
    \textbf{Chunhao Tian}\textsuperscript{1}\footnotemark[1],
    \textbf{Qi Qi}\textsuperscript{1}\thanks{Corresponding authors: Qi Qi and Hao Sun.},
    \textbf{Hao Sun}\textsuperscript{1}\footnotemark[2] \\
    \textsuperscript{1}Gaoling School of Artificial Intelligence,
    Renmin University of China, Beijing, China \\
    \texttt{\{myzhao13,tianchunhao,qi.qi,haosun\}@ruc.edu.cn}
}

\begin{document}
\maketitle

\begin{abstract}
Large language models (LLMs) can achieve strong reasoning performance with sufficient computation, but they do not inherently know how much computation a task requires. We study budgeted inference-time reasoning for multiple problems under a strict global token constraint and formulate it as an Ordered Stochastic Multiple-Choice Knapsack Problem (OS-MCKP). This perspective highlights a meta-cognitive requirement---anticipating problem difficulty and allocating computation strategically. We propose ROI-Reasoning, a two-stage framework that endows LLMs with intrinsic, budget-aware rationality. First, Meta-Cognitive Fine-Tuning teaches models to predict reasoning cost levels and decide whether a problem is worth attempting before generation. Next, Rationality-Aware Reinforcement Learning optimizes sequential decision making under a hard token budget, allowing models to learn long-horizon allocation strategies. Experiments across mathematical reasoning, multiple-choice question answering, and code generation show that ROI-Reasoning improves scores and generally reduces reference-order regret under tight budgets, with up to a \textbf{417\%} relative score improvement over the strongest same-backbone baseline in mathematics.

\end{abstract}
\section{Introduction}

People often face situations where they must plan ahead despite uncertainty about the cost and benefit of future actions. Consider Bob taking a timed exam. He can see all questions in advance but does not know how long each will take or whether he can solve it. Spending too long on a difficult question may leave easier ones unanswered; moving on too soon may sacrifice a solvable one. To maximize his score, Bob must decide how much time to spend on each question.

\begin{table}[t]
\centering
\footnotesize
\setlength{\tabcolsep}{1pt}
\renewcommand{\arraystretch}{1}
\begin{tabular}{@{}
  >{\raggedright\arraybackslash}m{.28\columnwidth}
  *{2}{>{\centering\arraybackslash}m{.15\columnwidth}}
  >{\centering\arraybackslash}m{.16\columnwidth}
  >{\centering\arraybackslash}m{.20\columnwidth}
  @{}}
\toprule
\textbf{Method}
& \shortstack{\textbf{Budget}\\\textbf{Control}}
& \shortstack{\textbf{Cost}\\\textbf{Estimate}}
& \shortstack{\textbf{Problem}\\\textbf{Skipping}}
& \shortstack{\textbf{Cross-Task}\\\textbf{Policy}} \\
\midrule
Self-Refine      & \roino  & \roino  & \roino  & \roino  \\
Budget Forcing   & \roiyes & \roino  & \roino  & \roino  \\
TALE-EP          & \roiyes & \roiyes & \roino  & \roino  \\
Plan-and-Budget  & \roiyes & \roiyes & \roino  & \roino  \\
\rowcolor{roiTint}
\textbf{ROI-Reasoning}
                 & \roiyes & \roiyes & \roiyes & \roiyes \\
\bottomrule
\end{tabular}
\caption{Capabilities relevant to reasoning across multiple problems under a shared token budget.}
\label{tab:method_capabilities}
\end{table}

\begin{figure*}[t]
    \centering
    \includegraphics[width=\textwidth]{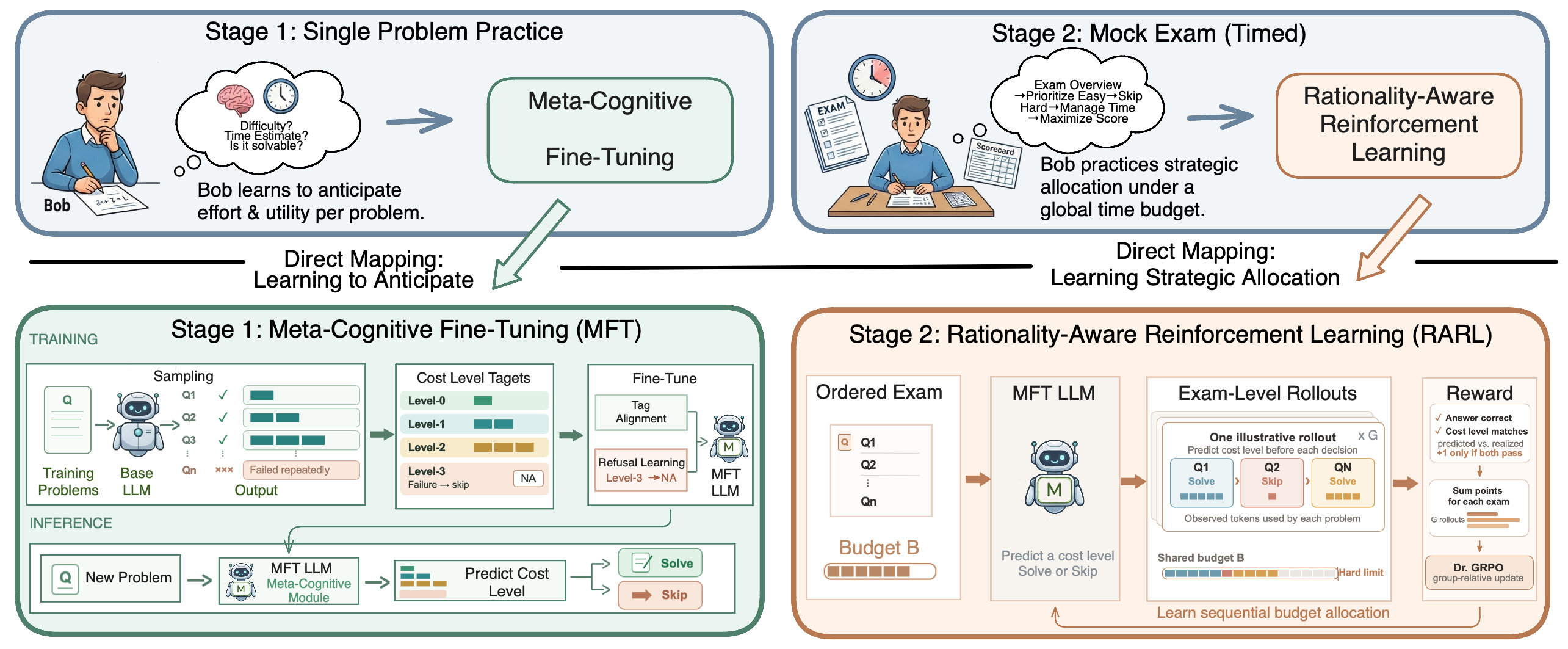}
    \caption{Overview of ROI-Reasoning. The top row illustrates the exam analogy; the bottom row shows MFT cost-level supervision and RARL policy optimization under a shared token budget.}
    \label{fig:roi-reasoning}
\end{figure*}

Large language model (LLM) agents face similar trade-offs in real-world applications, where users care about timely task completion and prefer lower token costs as long as the task is successfully completed. For example, a coding agent may need to review multiple pull requests under a limited usage budget. A routine review may require little analysis, whereas tracing a subtle bug across several files may consume far more. Because each review draws on the same budget, the agent needs \textbf{meta-cognitive planning}: estimating which changes warrant further analysis and when to stop or defer a review.

Existing approaches address only parts of this planning problem. Chain-of-Thought and self-reflection improve individual answers~\cite{wei2022chain,kojima2022large,shinn2023reflexion,madaan2023self}, while test-time scaling, including budget forcing, controls computation for a response~\cite{snell2024scaling,muennighoff2025s1}. TALE-EP~\cite{han2025token} estimates a per-question budget, and Plan-and-Budget~\cite{lin2026plan} allocates tokens among subquestions of a complex query. Table~\ref{tab:method_capabilities} highlights the gap: these methods neither learn a meta-cognitive policy for allocating a shared budget across distinct problems nor treat skipping a problem as a deliberate way to preserve computation for later ones.

To bridge this gap, we model large language model inference under a global token budget as an \emph{Ordered Stochastic Multiple-Choice Knapsack Problem} (OS-MCKP). In this formulation, each problem corresponds to a class, and different levels of reasoning effort correspond to items within that class, each associated with a computational cost (token usage) and a reward (answer quality). The total token budget serves as the knapsack capacity. The problem is \emph{stochastic} in that rewards are only revealed after generation, and \emph{ordered} in that problems must be processed sequentially, such that early decisions irreversibly constrain the remaining budget. 
This abstraction captures the core challenge of inference-time decision making in LLMs: allocating limited computation across problems to maximize overall performance under uncertainty.

Building on this formulation, we introduce \textbf{ROI-Reasoning}, a meta-cognitive reasoning framework that performs \textbf{R}ational \textbf{O}ptimization for \textbf{I}nference, where ROI naturally corresponds to \emph{Return on Investment}. 
Our approach consists of two complementary stages. We first introduce \textbf{Meta-Cognitive Fine-Tuning}, which leverages rejection sampling fine-tuning (RFT)~\cite{touvron2023llama,yuan2023scalingrelationshiplearningmathematical} to enable \emph{pre-computation}: before generating any solution, the model learns to anticipate problem difficulty, predict reasoning cost levels, and decide whether a problem is worth attempting under the available budget. We then propose \textbf{Rationality-Aware Reinforcement Learning}, which trains the model to plan and allocate computation across multiple problems under a hard global token constraint. 
Together, these two stages enable LLMs to plan ahead and adaptively allocate inference-time computation.

Our main contributions are summarized as follows. First, we formulate reasoning over an ordered sequence of problems under a shared token budget as OS-MCKP. Second, we propose\textbf{ ROI-Reasoning}, a two-stage framework for cost assessment and budget allocation. Third, we evaluate the framework on mathematical reasoning, multiple-choice question answering, and code generation, and conduct ablation studies and tests on unseen budgets and sequence lengths to assess its effectiveness and generalization.

\section{Related Work}

Chain-of-thought (CoT)~\cite{wei2022chain,kojima2022large} improves performance on multi-step reasoning tasks by eliciting intermediate steps. Self-Refine~\cite{madaan2023self} iteratively revises individual responses using model-generated feedback. Reinforcement-learning-based reasoning models~\cite{guo2025deepseek,openai2024learningtoreason,team2025kimi,qwen3} have also shown substantial gains in reasoning ability. At inference time, however, generating extended reasoning traces or repeatedly revising an answer can increase token use and computation. This has motivated work on efficient reasoning and explicit control of inference-time computation. Budget forcing~\cite{muennighoff2025s1} controls reasoning length by terminating or extending generation, while AdaptThink~\cite{zhang2025adaptthink} uses reinforcement learning to switch between thinking and no-thinking modes. Anytime reasoning~\cite{NEURIPS2025_21efcb6a} enables valid predictions even when the reasoning process is truncated. Adaptive computation controllers~\cite{huang2025adactrl} adjust reasoning effort to problem difficulty, and budget guidance~\cite{li2025steering} uses predictions of remaining reasoning length to steer generation. These methods primarily regulate computation within an individual response.

Beyond response-level control, some methods estimate reasoning costs or plan how tokens should be spent. TALE-EP~\cite{han2025token} estimates a suitable budget for a question before incorporating it into the answer prompt. Plan-and-Budget~\cite{lin2026plan} decomposes a complex query into subproblems and allocates tokens among them according to estimated complexity. Both account for reasoning effort, but their decisions concern one question or the subproblems of one query. In our setting, multiple independently scored problems are visible upfront, share a global budget, and must be answered in a fixed order. The model must therefore decide whether to attempt the current problem and how much computation to reserve for later ones as the remaining budget changes.

This shared-budget setting also connects to optimization under uncertainty. Predict-then-optimize~\cite{mandi2020smart,mulamba2020contrastive} estimates quantities such as costs or rewards before solving a downstream optimization problem. Reinforcement learning for combinatorial optimization~\cite{mazyavkina2021reinforcement,yang2023survey} provides another perspective on large discrete decision spaces with delayed rewards. By learning policies through interaction, it can address sequential decisions whose consequences cannot be reversed. These connections motivate our formulation of LLM reasoning as an ordered budget allocation problem with uncertain costs and rewards.

\section{Problem Formulation}
\label{sec:problem_formulation}

We study budgeted reasoning over multiple problems, motivated by scenarios such as answering an exam under a strict time limit. Let $\mathcal{X}$ denote a distribution over problem sequences. The model receives $N$ problems $\mathbf{x}=(x_1,\ldots,x_N)$ upfront in a single prompt but must answer them sequentially in a fixed order, without revisiting earlier decisions.

For each problem $x_i$, the model follows a policy $\pi_\theta(s_i,x_i)$, where $s_i$ summarizes the observable problem set, generation history, and remaining budget. An action may involve attempting a solution or strategically abstaining. It yields a non-negative reward $r(\pi_\theta(s_i,x_i))$, determined by a scoring function, and incurs a computational cost $c(\pi_\theta(s_i,x_i))$, measured in generated tokens. Neither the realized cost nor the reward is known in advance: token consumption is observed during generation, while reward is determined after evaluation.

The objective is to maximize expected total reward under a hard global token budget $B$:
\begin{equation}
\begin{aligned}
\max_{\pi_\theta} \quad
& \mathbb{E}_{\mathbf{x}\sim\mathcal{X},\,\pi_\theta}
\!\left[\sum_{i=1}^{N}r(\pi_\theta(s_i,x_i))\right] \\
\text{s.t.} \quad
& \sum_{i=1}^{N}c(\pi_\theta(s_i,x_i)) \leq B,
\end{aligned}
\end{equation}
where the expectation is over problem sequences and policy stochasticity, and the budget constraint applies to every rollout.

\paragraph{Knapsack Problems.}
This formulation generalizes the \emph{Multiple-Choice Knapsack Problem} (MCKP)~\cite{kellerer2004multiple}, where items are grouped into disjoint classes and at most one item is selected from each class. Each problem $x_i$ corresponds to a class, and alternative reasoning actions correspond to items within that class. Their rewards and token costs serve as item values and weights, respectively, while the global budget $B$ defines the knapsack capacity.

Unlike classical MCKP, which assumes deterministic values and weights, reasoning outcomes and costs are uncertain before execution. Both depend on the problem and the model's generation process, connecting our formulation to the \emph{Stochastic Multiple-Choice Knapsack Problem} (S-MCKP)~\cite{tran2015efficient}. The model must therefore allocate computation based on anticipated rather than known costs and rewards.

Our setting further imposes an explicit \emph{ordering constraint}. Although all problems are visible upfront, they must be processed in the given sequence. Spending computation on an early problem irreversibly reduces the budget available to later ones, so each decision must account for its opportunity cost across the remaining sequence.

Together, these properties define an \textbf{Ordered Stochastic Multiple-Choice Knapsack Problem (OS-MCKP)}. This perspective motivates our two-stage approach: learning to assess individual problems before reasoning, then learning to allocate computation across the full sequence under a shared budget.

\section{Methodology}

\subsection{Overview}

ROI-Reasoning separates per-problem assessment from sequence-level budget
allocation. MFT teaches cost prediction and solve-or-skip decisions
before reasoning, while RARL optimizes computation allocation across
the ordered sequence under a shared budget.
Figure~\ref{fig:roi-reasoning} summarizes the two-stage framework, connecting the exam analogy with the corresponding MFT and RARL workflows.

\subsection{Meta-Cognitive Fine-Tuning}
\label{subsec:meta-cognitive-fine-tuning}

Meta-Cognitive Fine-Tuning (MFT) equips the model with explicit cost
awareness before detailed reasoning. Rather than immediately attempting
a problem, the model first estimates its reasoning cost and decides
whether the expected benefit justifies that cost under the available
budget.

We make this assessment explicit through a cost-level tag placed before
the reasoning:
\begin{equation*}
\texttt{\fontsize{9}{9}\selectfont <predicted\_level>Level-}k%
\texttt{\fontsize{9}{9}\selectfont </predicted\_level>}
\end{equation*}
Here, $k$ is a coarse proxy for expected token consumption:
\texttt{Level-0}, \texttt{Level-1}, and \texttt{Level-2} respectively
denote solutions expected to require fewer than 256, 256--512, and more
than 512 tokens. \texttt{Level-3} indicates that the problem should be
skipped, in which case the model directly outputs
\texttt{\textbackslash boxed\{NA\}}.

MFT comprises two stages: \textbf{Tag Alignment} and
\textbf{Refusal Learning}. Tag Alignment uses rejection-sampling
fine-tuning to associate problems with cost levels calibrated from
empirical token usage. We first train on individual problems to learn
atomic cost assessment and the required output format, and then extend
the supervision to multi-problem sequences so that predictions account
for competing problems under a shared budget.

Refusal Learning introduces explicit supervision for strategic
abstention. We sample multiple solutions for each problem and label
persistent failures as \texttt{Level-3} with
\texttt{\textbackslash boxed\{NA\}}. Applying this supervision at both
the individual-problem and sequence levels teaches the model to avoid
low-ROI attempts and preserve computation for later problems. Dataset
construction, rejection-sampling rules, and intermediate training
stages are detailed in Appendix~\ref{app:meta-cognitive}.

\subsection{Rationality-Aware Reinforcement Learning}
\label{subsec:rl}

While MFT provides structured demonstrations of ROI-aware behavior, it remains imitation-based and cannot fully address the long-horizon credit assignment required for allocating a shared budget across multiple problems. We therefore introduce \textbf{Rationality-Aware Reinforcement Learning (RARL)}, which enables the model to learn budget-aware planning through interaction in simulated exam environments under an explicit global token budget $B$.

In each training sample, the model is presented with $N$ problems at once, forming a single exam instance. The problems are ordered and must be processed sequentially. A hard global constraint is imposed: the total number of generated tokens across the entire exam instance must not exceed $B$. This constraint is enforced in two ways. First, the budget $B$ is explicitly provided in the prompt, encouraging the model to reason with resource allocation across the problem sequence. Second, generation is terminated once the token limit is reached, ensuring that all sampled trajectories strictly satisfy the budget constraint. Under this formulation, the objective is to maximize the total return under a global budget constraint.

\begin{table*}[t]
\centering

{\small
\setlength{\tabcolsep}{3pt}
\renewcommand{\arraystretch}{1.0}

\begin{tabular}{lcccccccccc}
\toprule
\multirow{3}{*}{\textbf{Method}}
& \multicolumn{2}{c}{\textbf{Score$_{\mathrm{easy}}$}}
& \multicolumn{4}{c}{\textbf{Medium}}
& \multicolumn{4}{c}{\textbf{Hard}} \\
\cmidrule(lr){2-3}
\cmidrule(lr){4-7}
\cmidrule(lr){8-11}

& \multirow{2}{*}{1024}
& \multirow{2}{*}{512}
& \multicolumn{2}{c}{1024}
& \multicolumn{2}{c}{512}
& \multicolumn{2}{c}{1024}
& \multicolumn{2}{c}{512} \\
\cmidrule(lr){4-5}
\cmidrule(lr){6-7}
\cmidrule(lr){8-9}
\cmidrule(lr){10-11}

& &
& \textbf{Score}$\uparrow$ & \textbf{Reg.}$\downarrow$
& \textbf{Score}$\uparrow$ & \textbf{Reg.}$\downarrow$
& \textbf{Score}$\uparrow$ & \textbf{Reg.}$\downarrow$
& \textbf{Score}$\uparrow$ & \textbf{Reg.}$\downarrow$ \\
\midrule

\multicolumn{11}{c}{\textbf{Large-Scale Models}} \\

DeepSeek-V3.2 (685B)
& 1.85 & 1.53
& \underline{1.33} & 0.28
& \underline{0.88} & 0.42
& 0.87 & 0.53
& \underline{0.49} & 0.68 \\

GPT-4o-mini
& 1.53 & 1.19
& \textbf{1.41} & 0.08
& 0.65 & 0.45
& \textbf{1.27} & 0.17
& 0.32 & 0.73 \\

\midrule

\multicolumn{11}{c}{\textbf{Open-Source Models (1.5B--8B)}} \\

Llama-3.1-8B-Instruct
& 1.03 & 0.96
& 0.74 & 0.28
& 0.54 & 0.44
& 0.55 & 0.47
& 0.28 & 0.71 \\

Qwen2.5-Math-7B-Instruct
& 1.60 & 1.09
& 1.07 & 0.33
& 0.51 & 0.53
& 0.69 & 0.57
& 0.18 & 0.83 \\

Qwen2.5-7B-Instruct
& 1.55 & 1.24
& 1.27 & 0.18
& 0.64 & 0.48
& 1.07 & 0.31
& 0.29 & 0.77 \\

Llama-3.2-3B-Instruct
& 0.66 & 0.63
& 0.41 & 0.38
& 0.33 & 0.48
& 0.30 & 0.55
& 0.17 & 0.73 \\

Qwen2.5-Math-1.5B-Instruct
& 1.42 & 1.01
& 1.05 & 0.26
& 0.49 & 0.51
& 0.79 & 0.44
& 0.19 & 0.81 \\

\quad + Plan-and-Budget with Jev
& 1.43 & 1.01
& 1.08 & 0.24
& 0.50 & 0.50
& 0.86 & 0.40
& 0.20 & 0.80 \\

Qwen2.5-1.5B-Instruct
& 0.72 & 0.55
& 0.60 & 0.17
& 0.33 & 0.40
& 0.49 & 0.32
& 0.15 & 0.73 \\

\quad + Plan-and-Solve
& 0.61 & 0.35
& 0.40 & 0.34
& 0.15 & 0.57
& 0.28 & 0.53
& 0.05 & 0.85 \\

\quad + Least-to-Most
& 0.65 & 0.44
& 0.52 & 0.20
& 0.28 & 0.36
& 0.38 & 0.41
& 0.13 & 0.71 \\

\quad + Plan-and-Budget with Jev
& 0.64 & 0.58
& 0.51 & 0.19
& 0.31 & 0.47
& 0.41 & 0.36
& 0.16 & 0.73 \\

\quad + TALE-EP
& 0.45 & 0.39
& 0.43 & \underline{0.05}
& 0.28 & \underline{0.30}
& 0.38 & \underline{0.16}
& 0.18 & \underline{0.53} \\

\rowcolor[RGB]{236,244,252}
\quad \textbf{+ ROI-Reasoning}
& 1.15 & 1.08
& 1.13 & \textbf{0.02}
& \textbf{0.97} & \textbf{0.11}
& \underline{1.12} & \textbf{0.02}
& \textbf{0.93} & \textbf{0.14} \\
\bottomrule
\end{tabular}
}
\caption{
Main results under shared 1024- and 512-token budgets.
We report the easy-to-hard reference score
(Score$_{\mathrm{easy}}$), Score, and Reference-Order Regret.
Each test paper contains three problems and has a maximum score of 3.
For the Medium and Hard settings, the best and second-best values in
each Score and Regret column are shown in \textbf{bold} and
\underline{underlined}, respectively.
}
\label{tab:main_results}
\end{table*}

\paragraph{Calibrated surrogate reward.}

RARL optimizes a calibrated surrogate reward that couples answer correctness with cost-level calibration.

This design induces a natural \emph{meta-cognitive self-reflection} mechanism: before committing to detailed reasoning, the model is incentivized to assess how much computation a problem is likely to require and whether that investment is justified under the remaining budget. 

For problem $i$, let $\mathrm{Tok}_i$ denote the actual number of generated tokens consumed for that problem, and let $\ell_i = \mathrm{Level}(\mathrm{Tok}_i)$ be the posterior cost level induced by token consumption. Let $\hat{\ell}_i$ be the cost level predicted by the model before reasoning. The per-problem reward is defined as
\begin{equation}
r_i
=
\mathbb{I}\!\left[
\mathrm{correct}_i
\wedge
\hat{\ell}_i = \ell_i
\right].
\end{equation}
The exam-level reward is then $R=\sum_{i=1}^{N} r_i,$ which is bounded above by $N$.

\paragraph{Policy Optimization.}
We adopt Dr.~GRPO~\cite{shao2024deepseekmath,liu2025understanding} because standard GRPO normalizes updates by response
length and group reward standard deviation, which can introduce
length and problem-difficulty biases. Such biases are particularly
problematic in our setting, where reasoning length is itself an
allocation decision. By removing both normalization terms, Dr.~GRPO
ensures that policy updates are driven by the calibrated exam-level
reward rather than optimizer-induced preferences.

For each test paper $q$, we sample $G$ complete rollouts
$\{o_g\}_{g=1}^{G}$ from the old policy
$\pi_{\theta_{\mathrm{old}}}$ and maximize
\begin{equation}
\begin{aligned}
J(\theta)
&=
\mathbb{E}\Bigg[
\frac{1}{G}
\sum_{g=1}^{G}
\sum_{t=1}^{|o_g|}
\min\Big(
\rho_{g,t}(\theta)\widehat{A}_g,\\
&\qquad
\operatorname{clip}\!\left(
\rho_{g,t}(\theta),1-\epsilon,1+\epsilon
\right)\widehat{A}_g
\Big)
\Bigg].
\end{aligned}
\label{eq:rarl_objective}
\end{equation}
Here, the token-level importance ratio is
\begin{equation}
\rho_{g,t}(\theta)
=
\frac{
\pi_\theta(o_{g,t}\mid q,o_{g,<t})
}{
\pi_{\theta_{\mathrm{old}}}(o_{g,t}\mid q,o_{g,<t})
},
\label{eq:rarl_ratio}
\end{equation}
and all tokens in rollout $o_g$ share the group-relative advantage
\begin{equation}
\widehat{A}_g
=
R_g-\frac{1}{G}\sum_{j=1}^{G}R_j,
\label{eq:rarl_advantage}
\end{equation}
where $R_g$ is its calibrated exam-level reward and $\epsilon$ is
the clipping coefficient. Following Dr.~GRPO, we use neither
response-length normalization nor group reward standard-deviation
normalization.

\begin{figure*}[t]
    \centering
    \begin{subfigure}{0.48\textwidth}
        \centering
        \includegraphics[width=\linewidth]{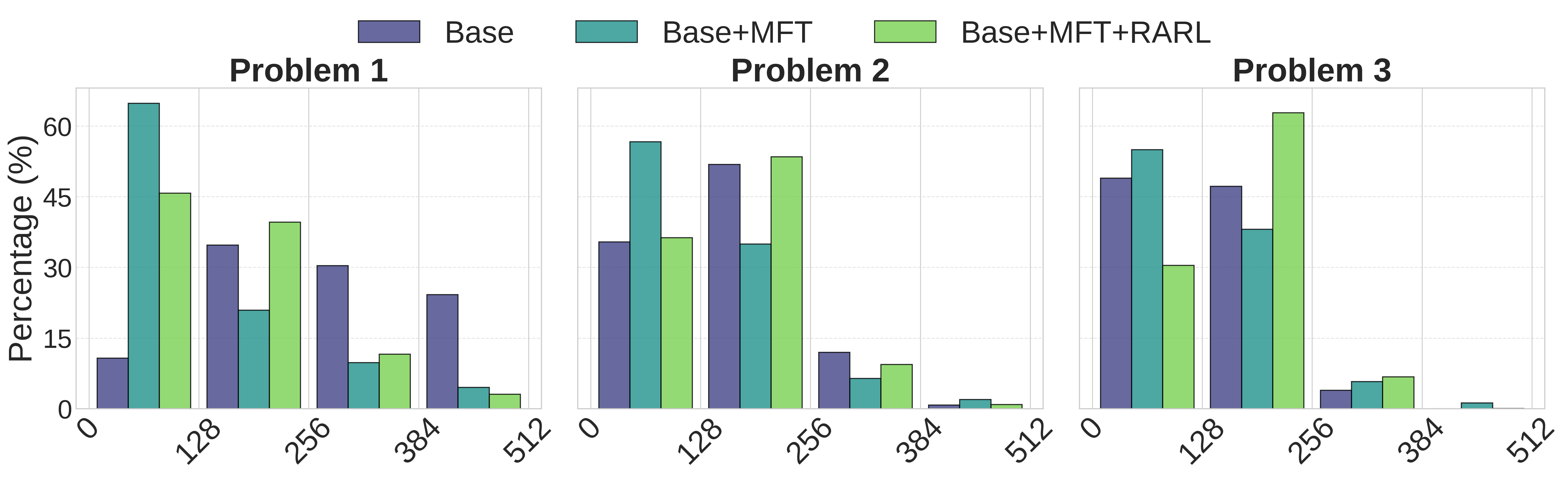}
        \caption{Medium with 512 Budget}
        \label{fig:dist_med_512}
    \end{subfigure}
    \hfill
    \begin{subfigure}{0.49\textwidth}
        \centering
        \includegraphics[width=\linewidth]{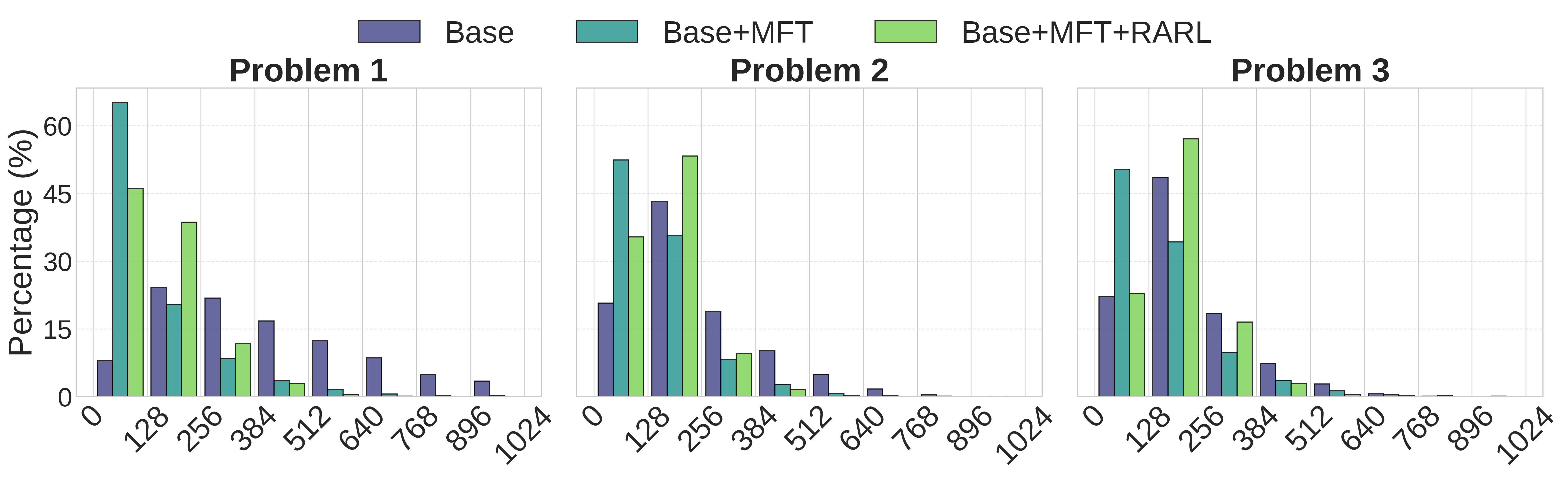}
        \caption{Medium with 1024 Budget}
        \label{fig:dist_med_1024}
    \end{subfigure}
    
    \vspace{1em} 
 
    \begin{subfigure}{0.49\textwidth}
        \centering
  
        \includegraphics[width=\linewidth]{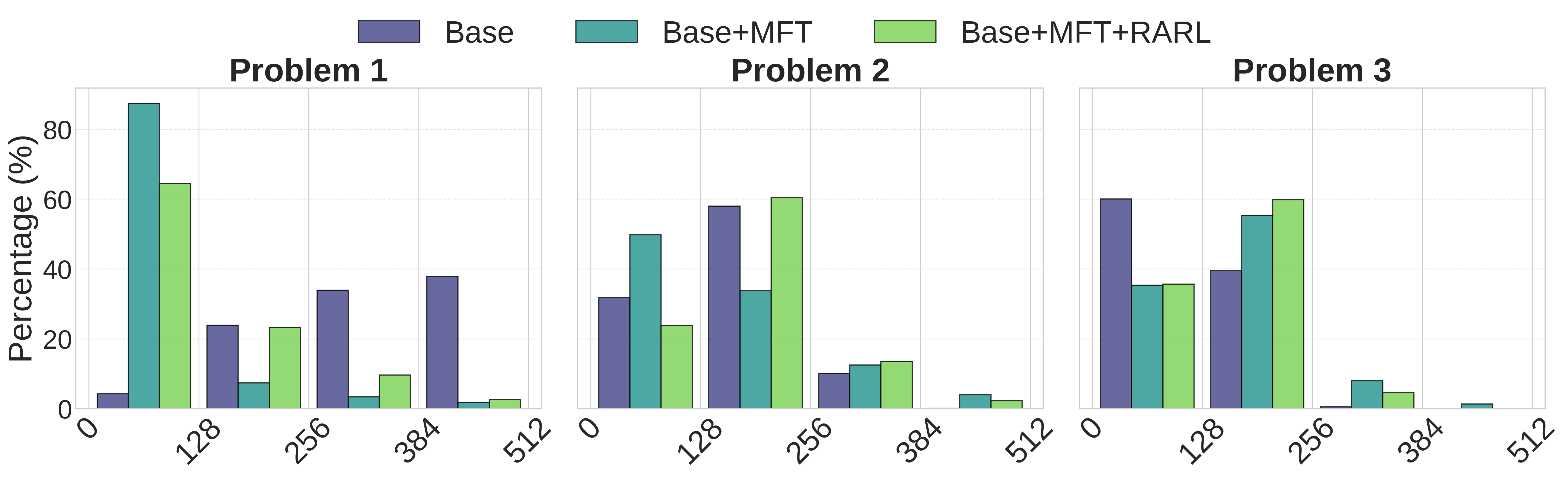}
        \caption{Hard with 512 Budget}
        \label{fig:dist_hard_512}
    \end{subfigure}
    \hfill
    \begin{subfigure}{0.48\textwidth}
        \centering
       
        \includegraphics[width=\linewidth]{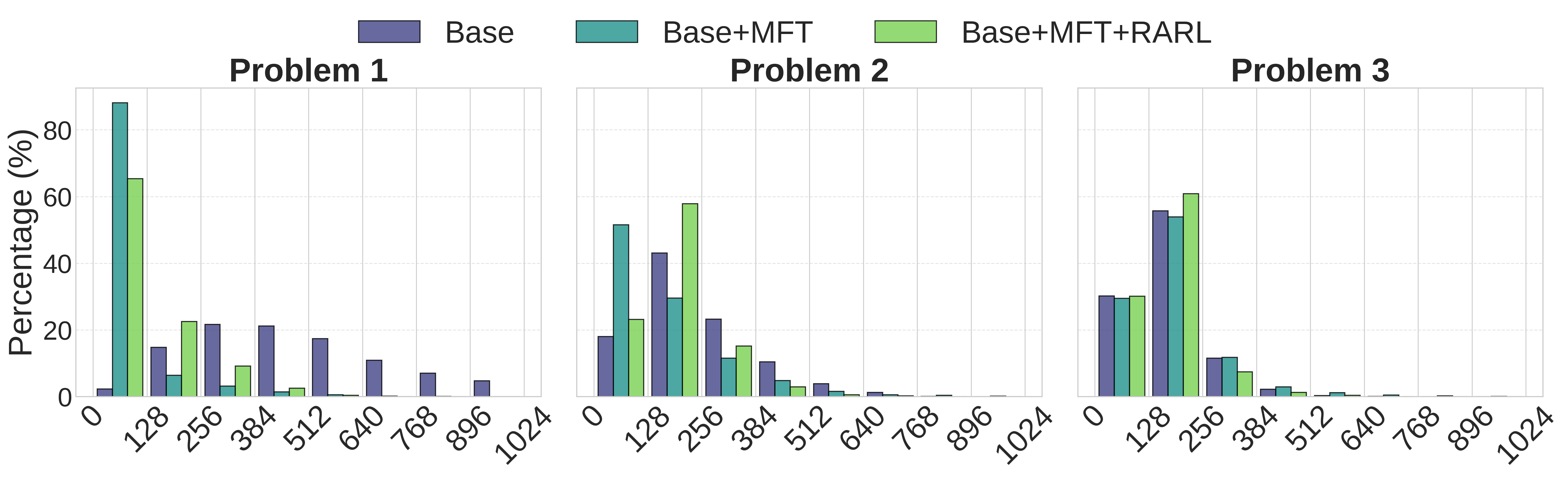}
        \caption{Hard with 1024 Budget}
        \label{fig:dist_hard_1024}
    \end{subfigure}

    \caption{Token-length distributions by problem position under different difficulty levels and token budgets. 
    The histograms compare the reasoning length patterns of \textbf{Qwen2.5-1.5B-Instruct}, \textbf{Qwen2.5-1.5B-Instruct + MFT}, and \textbf{Qwen2.5-1.5B-Instruct + MFT + RARL} under four distinct settings.}
    \label{fig:ablation_token_grid}
\end{figure*}

\begin{table*}[t]
\centering
\begingroup
\small
\setlength{\tabcolsep}{2.5pt}
\renewcommand{\arraystretch}{1.0}
\resizebox{0.9\textwidth}{!}{%
\begin{tabular}{@{}lcc@{\hspace{5pt}}cccccccc@{}}
\toprule
\multirow{3}{*}{\textbf{Variant}}
& \multirow{3}{*}{\textbf{Stage 1}}
& \multirow{3}{*}{\textbf{Stage 2}}
& \multicolumn{4}{c}{\textbf{Medium}}
& \multicolumn{4}{c}{\textbf{Hard}} \\
\cmidrule(lr){4-7} \cmidrule(lr){8-11}
& & &
  \multicolumn{2}{c}{1024}
& \multicolumn{2}{c}{512}
& \multicolumn{2}{c}{1024}
& \multicolumn{2}{c}{512} \\
\cmidrule(lr){4-5} \cmidrule(lr){6-7}
\cmidrule(lr){8-9} \cmidrule(lr){10-11}
& & &
  \textbf{Score}$\uparrow$ & \textbf{Reg.}$\downarrow$
& \textbf{Score}$\uparrow$ & \textbf{Reg.}$\downarrow$
& \textbf{Score}$\uparrow$ & \textbf{Reg.}$\downarrow$
& \textbf{Score}$\uparrow$ & \textbf{Reg.}$\downarrow$ \\
\midrule
\textbf{ROI-Reasoning}
& MFT & RARL
& 1.13 & \underline{0.02}
& \textbf{0.97} & 0.11
& 1.12 & \textbf{0.02}
& \textbf{0.93} & 0.14 \\

\addlinespace[2pt]
\quad w/o RARL
& MFT & --
& 0.98 & 0.03
& 0.82 & \underline{0.09}
& 0.95 & 0.04
& 0.81 & \underline{0.10} \\
\quad w/o RARL, w/ Greedy
& MFT & Greedy
& 0.97 & 0.04
& \underline{0.83} & 0.15
& 0.96 & 0.04
& \underline{0.83} & 0.15 \\

\addlinespace[2pt]
\quad w/o Cost Tags
& RFT & Dr.~GRPO
& \textbf{1.21} & \textbf{0.01}
& 0.80 & 0.20
& \textbf{1.19} & \underline{0.03}
& 0.68 & 0.32 \\
\quad w/o Cost Tags, RL only
& -- & Dr.~GRPO
& \underline{1.18} & 0.03
& 0.72 & 0.30
& \underline{1.15} & 0.04
& 0.51 & 0.50 \\
\quad w/o Cost Tags, FT only
& RFT & --
& 0.86 & 0.05
& 0.78 & \textbf{0.06}
& 0.85 & 0.06
& 0.76 & \textbf{0.08} \\

\midrule
Base
& -- & --
& 0.60 & 0.17
& 0.33 & 0.40
& 0.49 & 0.32
& 0.15 & 0.73 \\
\bottomrule
\end{tabular}%
}
\caption{Component ablation on Qwen2.5-1.5B-Instruct.
MFT supervises cost levels; RARL rewards correct answers with
cost-level agreement. RFT and Dr.~GRPO use
correctness-only objectives, while Greedy replaces RARL with
external allocation. Bold and underlined values indicate the
best and second-best results per column.}
\label{tab:component_ablation}
\endgroup
\end{table*}

\section{Experiments}

We evaluate ROI-Reasoning in three budgeted reasoning domains:
mathematical reasoning, multiple-choice question answering (MCQA),
and code generation. We focus on mathematical reasoning and provide
a compact cross-domain evaluation, with datasets, baselines,
experimental details, and full results deferred to the appendix. Our experiments examine whether MFT and RARL improve
computation allocation across multiple problems under a shared token
budget, beyond gains in single-problem accuracy.

\subsection{Experimental Setup}
\paragraph{Data.}
The training data is drawn from the GSM8K~\cite{cobbe2021gsm8k} and MATH~\cite{hendrycksmath2021} training sets alongside AIME problems from 2022--2024. Evaluation is conducted on the corresponding GSM8K and MATH test sets, as well as AIME25~\cite{aime25}.

\paragraph{Test Paper Construction and Budgets.}
Instead of evaluating problems in isolation, we group three problems into a single \emph{test paper}, which constitutes one query to the model, and construct 5,000 underlying problem sets, each instantiated as a paired Medium and Hard test paper. We assume an expected difficulty ordering of AIME $>$ MATH $>$ GSM8K, and construct each test paper by sampling three problems according to dataset templates and arranging them in different orders. We consider two settings: \emph{Medium} papers, which interleave problems of different difficulty, and \emph{Hard} papers, which place more challenging problems earlier in the sequence. For the main $N=3$ evaluation, Medium and Hard papers contain the same underlying problem sets but differ in their presentation order; therefore, they share the same easy-to-hard reference score. Detailed construction procedures are provided in Appendix~\ref{app:test_paper}.

Training uses $B=1024$, while evaluation uses $B\in\{1024,512\}$.
The former typically permits two full solutions, inducing
allocation trade-offs, whereas the latter makes early stopping critical.

\paragraph{Training Details.}
All trainable variants in the mathematical-reasoning experiments use Qwen2.5-1.5B-Instruct. MFT uses AdamW for three epochs, with a learning rate of $5\times10^{-6}$ and an effective batch size of 64. RARL uses AdamW with a learning rate of $1\times10^{-6}$, batch size 128, and $G=12$ rollouts per test paper. Further training settings are provided in Appendix~\ref{app:experimental_setup}.

\paragraph{Models and Baselines.}
We compare against frontier models, including GPT-4o-mini
\cite{hurst2024gpt} and DeepSeek-V3.2
\cite{deepseekai2025deepseekv32}, and open-source models from the
Qwen2.5~\cite{qwen2025qwen25technicalreport} and
Llama~3~\cite{grattafiori2024llama} families. All non-trained
baselines receive explicit budget-allocation instructions.
Same-backbone training-free baselines include
Plan-and-Solve~\cite{wang2023plan},
Least-to-Most~\cite{zhou2022least},
a Plan-and-Budget-inspired adaptation~\cite{lin2026plan}
using Jev~\cite{jevmodels} to predict per-problem cost and skip tags,
and a level-tag adaptation of TALE-EP~\cite{han2025token}
using the solver itself to predict tags before answering.
Details are provided in Appendix~\ref{app:baselines}.

\subsection{Evaluation Metrics}

We report \textbf{Score}, the mean total score per test paper.
To assess sensitivity to problem order under a shared token budget,
we also report \textbf{Reference-Order Regret} relative to a fixed
easy-to-hard ordering of the same problems. This reference
prioritizes lower-difficulty problems and is specified before
evaluating any model. Let $S_B(\pi)$ and $S_B^{\mathrm{easy}}(\pi)$
denote the mean scores of policy $\pi$ on the same underlying
problem sets in the original and easy-to-hard orders, respectively,
under budget $B$. We define
\begin{equation}
\mathrm{Reg}_B(\pi)
=
\frac{S_B^{\mathrm{easy}}(\pi)-S_B(\pi)}
     {S_B^{\mathrm{easy}}(\pi)}.
\end{equation}
The measure is defined when $S_B^{\mathrm{easy}}(\pi)>0$.

Lower values indicate better original-order performance relative to the
reference. Because the reference is fixed rather than selected
to maximize each policy's score, negative values are possible:
they indicate that the original order scores higher. Signed
values are discussed further in Appendix~\ref{app:mcq}.
Main results assign one point per correct answer;
heterogeneous-scoring results are provided in Appendix
Section~\ref{app:heterogeneous-reward}.

\begin{figure}[t]
    \centering
    \includegraphics[width=\linewidth]{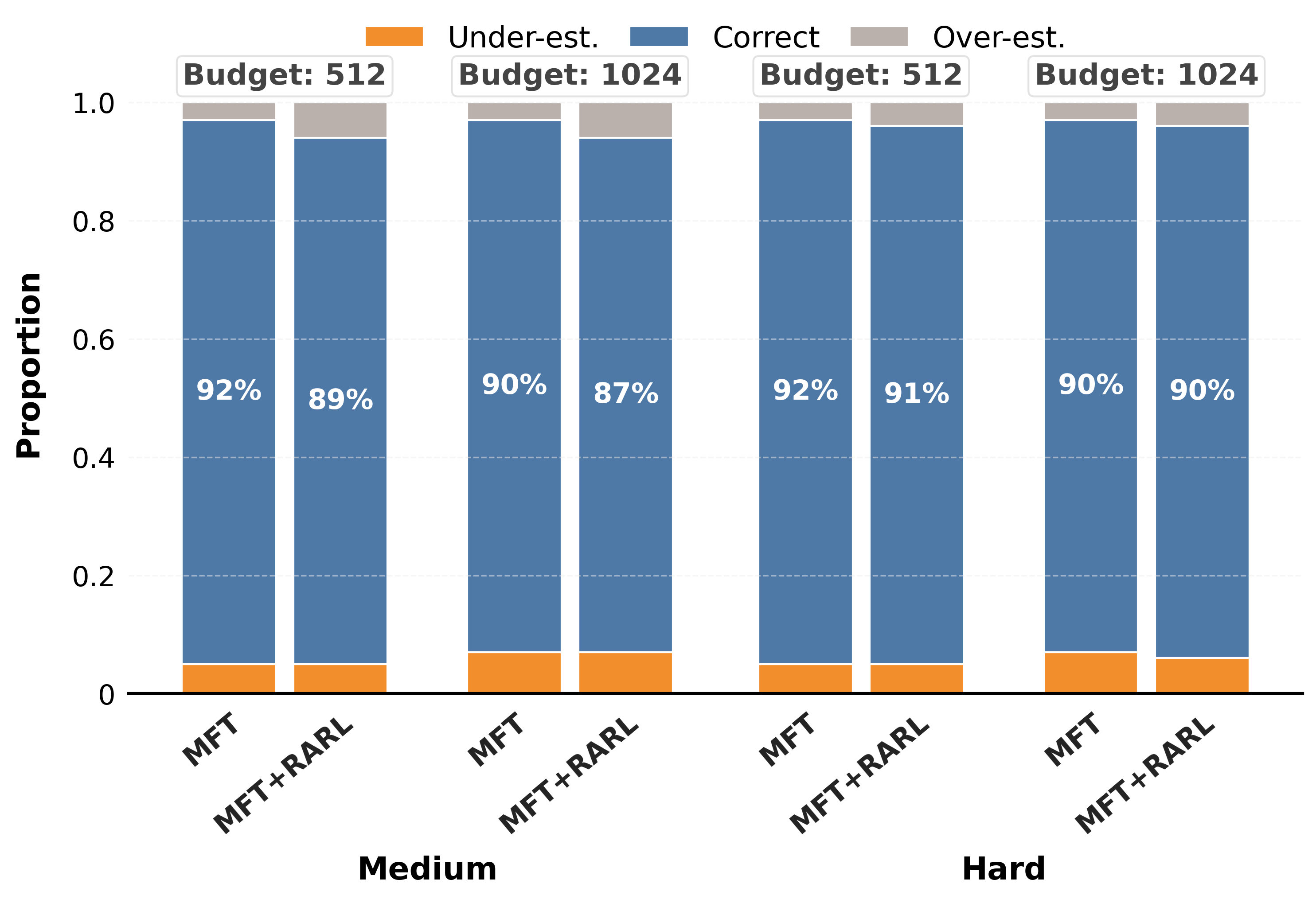}
    \caption{
Composition of predicted-level outcomes (correct, under-, and over-estimated) under different budgets on Medium and Hard test papers, comparing MFT and MFT+RARL.
}
    \label{fig:self_reflection}
\end{figure}

\begin{figure*}[t]
    \centering
    \includegraphics[width=\textwidth]{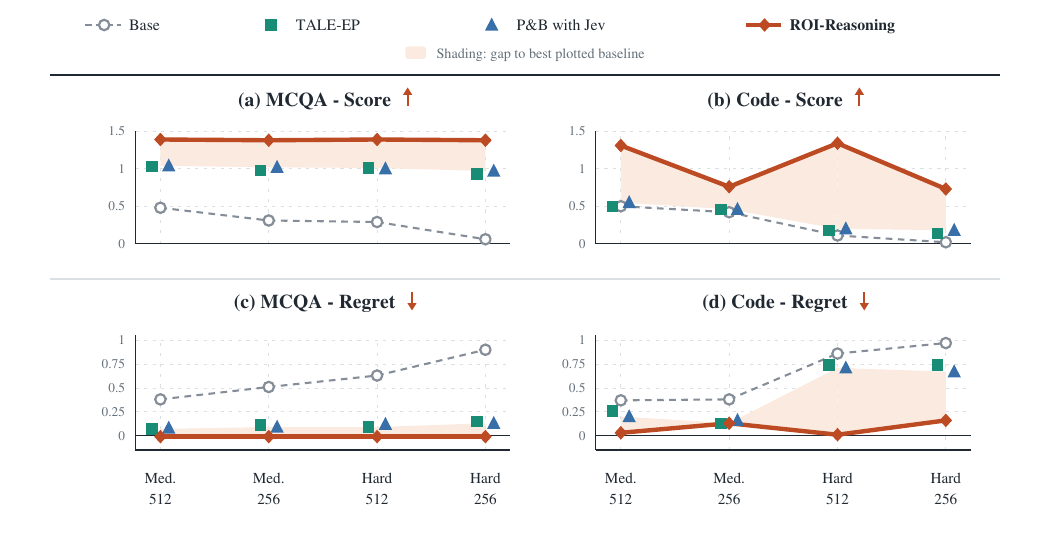}
\caption{Cross-domain Score and Regret for same-backbone methods on MCQA and code generation under shared 512- and 256-token budgets.}
    \label{fig:cross_domain_results}
\end{figure*}

\subsection{Main Results}

Table~\ref{tab:main_results} shows that ROI-Reasoning improves both Score and reference-order Regret over the same-backbone base model and training-free baselines in all four settings. At 512 tokens, it achieves the highest Score and lowest Regret in the entire table; at 1024 tokens, it retains the lowest Regret with competitive Scores. Among same-backbone baselines, TALE-EP reduces Regret but often lowers Score, while Plan-and-Budget with Jev does not consistently improve either metric over the base model. ROI-Reasoning also outperforms a mathematics-specialized model paired with a general-purpose planner in both metrics across all settings. On Hard-512 papers, their Scores are 0.93 and 0.20, respectively, with Regret of 0.14 and 0.80.

These results further show that frontier-model status and parameter scale do not reliably predict performance under a shared reasoning budget. GPT-4o-mini and DeepSeek-V3.2 do not maintain an advantage across settings, while Qwen2.5-Math-7B-Instruct does not consistently outperform its 1.5B counterpart or Qwen2.5-7B-Instruct. Strong general or mathematical reasoning ability therefore does not by itself ensure effective allocation across problems. Although some baselines are competitive in individual settings, ROI-Reasoning delivers more consistent gains in both Score and Regret.

\begin{table}[t]
\centering
{\small
\setlength{\tabcolsep}{5pt}
\renewcommand{\arraystretch}{1.02}
\resizebox{0.95\columnwidth}{!}{%
\begin{tabular}{@{}lccc@{}}
\toprule
\multirow{2}{*}{\textbf{$N/B$}}
& \multicolumn{3}{c}{\textbf{Score (\%, Medium / Hard)}} \\
\cmidrule(lr){2-4}
& \textbf{Base}
& \textbf{MFT}
& \textbf{Ours} \\
\midrule

\multicolumn{4}{c}{\textit{Unseen $N$}} \\

$4/1024$
& 18.66/14.87
& 32.91/30.37
& \textbf{39.49/37.23} \\

$4/1536$
& 22.72/19.91
& 33.86/31.36
& \textbf{39.85/37.50} \\

$5/1024$
& 16.37/12.64
& 31.71/27.82
& \textbf{40.56/35.39} \\

$5/1536$
& 20.66/17.80
& 33.29/29.10
& \textbf{41.04/36.16} \\

\midrule

\multicolumn{4}{c}{\textit{Loose $B$}} \\

$3/2048$
& 23.09/21.35
& 31.31/30.69
& \textbf{38.05/37.55} \\

$3/4096$
& 22.57/21.27
& 31.67/30.85
& \textbf{38.00/37.69} \\

\bottomrule
\end{tabular}
}}
\caption{
Generalization to unseen sequence lengths and token budgets under
uniform per-problem scoring. $N/B$ denotes sequence length/token
budget, and normalized scores are reported as Medium/Hard.
}
\label{tab:generalization}
\end{table}

\subsection{Additional Evaluations and Analysis}
\paragraph{Ablation Study.}
Table~\ref{tab:component_ablation} shows that MFT improves Score and Regret over the base model across all settings. RARL adds 0.12--0.17 Score, while Greedy provides little additional benefit. At the tighter, unseen 512-token budget, ROI-Reasoning exceeds the tag-free two-stage variant by 0.17/0.25 Score on Medium/Hard papers, with lower Regret. At the seen 1024-token budget, its Scores are modestly lower by 0.08/0.07. RL only scores below the tag-free two-stage variant; FT only achieves lower 512-token Regret than ROI-Reasoning but also lower Scores. Together, these results support the complementary benefits of MFT and RARL and the value of the complete cost-aware design for robustness to budget contraction.

\paragraph{Token Allocation and Cost-Level Prediction.}
Figure~\ref{fig:ablation_token_grid} shows that the base model concentrates token use on earlier problems, whereas MFT and MFT+RARL (ROI-Reasoning) preserve more of the shared budget for later ones. Figure~\ref{fig:self_reflection} shows 87--92\% agreement between predicted and realized cost levels across the evaluated settings. Adding RARL preserves this agreement while improving allocation across the paper. Appendix~\ref{app:per_problem_accuracy} reports the corresponding per-problem accuracy, and Appendix~\ref{app:cost_underestimation_case} examines a remaining failure case in which cost underestimation causes truncation before the final problem can be answered.

\paragraph{Cross-Domain Applicability.}
We adapt ROI-Reasoning to multiple-choice question answering (MCQA) and code generation. Unlike in mathematical reasoning, TALE-EP and Plan-and-Budget with Jev generally improve Score and consistently reduce reference-order Regret over the base model in these domains. ROI-Reasoning improves further, achieving the highest Score and lowest Regret among the same-backbone methods shown in Figure~\ref{fig:cross_domain_results} across both budgets and paper settings. The advantage is especially pronounced on Code Hard-256 papers, where success requires executable code rather than a short MCQA option. Compared with Plan-and-Budget with Jev, ROI-Reasoning raises Score by 306\% (0.18 to 0.73) and reduces Regret by 76\% (0.67 to 0.16).
Full results and experimental details are provided in Appendix~\ref{app:generalization}.

\paragraph{Generalization to Unseen Sequence Lengths and Budgets.}
Table~\ref{tab:generalization} evaluates unseen sequence lengths ($N=4,5$) under 1024- and 1536-token budgets, as well as larger unseen budgets ($B=2048,4096$) at $N=3$. In every setting, MFT improves on the base model, and ROI-Reasoning further improves on MFT for both Medium and Hard papers, by 5.99--8.85 percentage points. These results show that the learned allocation policy generalizes beyond the training configuration of $N=3$ and $B=1024$, while retaining its advantage when more computation is available. Full results on unseen sequence lengths appear in Appendix ~\ref{app:unseen_sequence_lengths}.

\section{Conclusion}

We studied budgeted inference-time reasoning across multiple problems under a strict global token constraint and formulated it as an OS-MCKP, capturing uncertain rewards and the consequences of a fixed problem order.

We proposed ROI-Reasoning, which combines MFT for cost-level prediction and solve-or-skip decisions with RARL for policy optimization. Across mathematical reasoning, MCQA, and code generation, it achieves higher Scores and generally lower Regret than same-backbone baselines, especially under tight budgets. Ablations and tests on unseen sequence lengths and budgets show that both stages contribute to the gains and that these gains extend beyond the training setting.

\section*{Limitations}

ROI-Reasoning assumes that all problems are visible before generation
and that answers are produced in a fixed order under a shared
output-token budget. The discrete cost levels provide only a coarse
proxy for computation and do not directly capture latency, tool use,
verification cost, or memory. Moreover, \texttt{Level-3} is treated as
a hard skip rather than a low-cost probe followed by reassessment.
Although the experiments cover mathematics, MCQA, code generation,
unseen sequence lengths, and unseen budgets, they do not address
dependencies between problems or unconstrained single-problem
reasoning. These settings remain important directions for future work.

\bibliography{references}

\clearpage
\appendix
\section{Meta-Cognitive Fine-Tuning}
\label{app:meta-cognitive}

This appendix provides implementation details of Meta-Cognitive Fine-Tuning (MFT). The procedure is designed to enhance the model's meta-cognitive capabilities through two stages: \textit{Tag Alignment} and \textit{Refusal Learning}. Training alternates between single-problem instances and short multi-problem sequences to progressively bridge atomic cost estimation and contextual decision making.

\subsection*{Stage I: Tag Alignment}

The goal of Tag Alignment is to train the model to correctly predict cost levels using the 

\texttt{<predicted\_level>} tag and to adhere to the standardized reasoning template.

\paragraph{Single-Problem Adaptation.}
We first construct a dataset of single-problem instances from GSM8K and MATH. Starting from the base model, we apply Rejection Sampling Fine-Tuning (RFT) and retain only samples that both produce correct final answers and strictly adhere to the required output format, including a valid \texttt{<predicted\_level>} tag. For retained samples, difficulty tags are calibrated based on empirical token consumption and difficulty statistics. The resulting model is denoted as $\mathcal{M}_{\text{1-Q-v1}}$.

\paragraph{Sequential Adaptation.}
To expose the model to contextual budgeting, we construct three-problem sequences by concatenating GSM8K and MATH instances. Initializing from $\mathcal{M}_{\text{1-Q-v1}}$, we perform RFT and retain sequences where all three problems are solved correctly and the \texttt{<predicted\_level>} tags follow the required format. This produces $\mathcal{M}_{\text{3-Q-v1}}$.

\subsection*{Stage II: Refusal Learning}

Refusal Learning introduces explicit supervision for strategic abstention by allowing the model to output a standardized \texttt{NA} token when problems are consistently unsolved.

\paragraph{Single-Problem Refusal.}
We extend the dataset to include harder problems from AIME 2022--2024. Starting from $\mathcal{M}_{\text{3-Q-v1}}$, we perform RFT with $K$ sampled generations per problem. If none of the samples yield a correct answer, the target output is labeled as:
\begin{center}
\fontsize{8pt}{8pt}\selectfont
\texttt{<predicted\_level>Level-3</predicted\_level> \textbackslash boxed\{\colorbox{gray!20}{NA}\}}
\end{center}
Otherwise, correct samples are retained and their difficulty tags are recalibrated. The resulting model is denoted as $\mathcal{M}_{\text{1-Q-v2}}$.

\paragraph{Sequential Refusal.}
Finally, we construct three-problem sequences drawn from GSM8K, MATH, and AIME. Initializing from $\mathcal{M}_{\text{1-Q-v2}}$, we perform RFT with $K$ samples per sequence. For fully correct sequences, difficulty tags are calibrated as before. For sequences containing persistent failures, the corresponding problem is labeled as:
\begin{center}
\fontsize{8pt}{8pt}\selectfont
\texttt{<predicted\_level>Level-3</predicted\_level> \textbackslash boxed\{\colorbox{gray!20}{NA}\}}
\end{center}
The final model, $\mathcal{M}_{\text{3-Q-v2}}$, serves as our Meta-Cognitive Fine-Tuned model.

\section{Main-Experiment Setup}
\label{app:experimental_setup}

We report the configuration used for the main mathematical-reasoning
experiment. Run-specific checkpoint paths, logging frequencies, and
memory-allocation flags are omitted because they do not change the
training objective or evaluation protocol.

\begin{table}[t]
\centering
\small
\begin{tabular}{@{}p{0.40\linewidth}p{0.50\linewidth}@{}}
\toprule
\textbf{MFT setting} & \textbf{Value} \\
\midrule
Backbone & Qwen2.5-1.5B-Instruct \\
Training & full-parameter SFT with LLaMA-Factory \\
Hardware & $4\times$ NVIDIA GeForce RTX 4090 \\
Precision & \texttt{bf16} \\
Optimizer & AdamW \\
Learning rate & $5\times10^{-6}$ \\
LR schedule / warmup & cosine / 0.03 \\
Training epochs & 3 \\
Cutoff length & 2048 \\
Per-device batch / accumulation & 2 / 8 \\
Effective batch size & 64 \\
Maximum gradient norm & 1.0 \\
Gradient checkpointing & enabled \\
Distributed training & DeepSpeed ZeRO-2 \\
\bottomrule
\end{tabular}
\caption{Main-experiment MFT configuration.}
\label{tab:mft_hyperparameters}
\end{table}

Table~\ref{tab:mft_hyperparameters} summarizes the supervised
meta-cognitive adaptation stage. The effective batch size is computed
as $2\times8\times4=64$ across the four RTX 4090 GPUs. The same
configuration is used to learn both the structured cost tags and the
refusal behavior described above.

\begin{table}[t]
\centering
\small
\begin{tabular}{@{}p{0.40\linewidth}p{0.50\linewidth}@{}}
\toprule
\textbf{RARL setting} & \textbf{Value} \\
\midrule
Initialization & final MFT checkpoint \\
Framework / rollout engine & \texttt{verl} / vLLM \\
Hardware & $4\times$ NVIDIA RTX PRO 6000 (one node) \\
Advantage estimator & Dr.~GRPO\\
KL regularization / entropy coefficient & none / 0.0 \\
PPO clipping coefficient / epochs & 0.2 / 1 \\
Optimizer & AdamW \\
Learning rate / schedule & $1\times10^{-6}$ / constant \\
Weight decay / AdamW betas & 0.01 / $(0.9,0.999)$ \\
Train batch size & 128 \\
Rollouts per instance $G$ & 12 \\
Temperature / top-$p$ & 1.0 / 1.0 \\
Maximum prompt / response length & 4096 / 1024 \\
Training paper length / budget & $N=3$ / $B=1024$ \\
Validation decoding & greedy ($n=1$) \\
\bottomrule
\end{tabular}
\caption{Main-experiment RARL configuration.}
\label{tab:rarl_hyperparameters}
\end{table}

As shown in Table~\ref{tab:rarl_hyperparameters}, RARL is initialized
from MFT and optimizes the complete three-problem paper under the same
1024-token budget used during training. Dr.~GRPO removes
group-standard-deviation normalization, while $G=12$ rollouts provide
within-instance comparisons. The rule-based reward couples answer
correctness with cost-level calibration, matching the objective
described in the main paper.

\section{Test Paper Construction}
\label{app:test_paper}

In this section, we describe how test papers are constructed from individual problems.
Each test paper consists of exactly three mathematics problems and is treated as a single query to the model.
The instruction text specifies a strict global token budget and a standardized answer format; since different baselines use different instructions, we focus here on the construction of problem sets and their ordering.

\paragraph{Problem Sources.}
Each test paper is formed by sampling problems from three standard datasets with increasing expected difficulty:
GSM8K (easy), MATH (medium), and AIME (hard).
All problems are drawn from the official test splits of the corresponding datasets.
For each dataset, we shuffle the problem pool once and then cycle through it sequentially to ensure balanced coverage across test papers.

\paragraph{Paper Length and Composition.}
We first construct 5{,}000 underlying problem sets, each containing
one problem from GSM8K, MATH, and AIME. For each set, we create paired
Medium and Hard papers using the same three problems but different
presentation orders. Thus, the two settings differ only in ordering,
not in problem content, and share the same easy-to-hard reference
paper.
\paragraph{Medium Test Papers.}
Medium papers interleave the three difficulty levels using predefined
ordering templates, such as (MATH, GSM8K, AIME) and
(GSM8K, AIME, MATH). A template is selected uniformly at random for
each underlying problem set. Across the templates, the hardest problem
is not systematically placed first.

\paragraph{Hard Test Papers.}
For each underlying problem set, the corresponding Hard paper places
more challenging problems earlier, using templates such as
(AIME, MATH, GSM8K) and (AIME, GSM8K, MATH). This ordering increases
the opportunity cost of early reasoning: excessive computation on the
first problem may leave insufficient budget for later problems.

\paragraph{Easy Variants and \texorpdfstring{$\text{score}_{\text{easy}}$}{score\_easy}.}
For each test paper, we additionally construct an \emph{easy} variant using the same set of three problems but a different ordering.
Specifically, problems are reordered from easiest to hardest according to the fixed priority
GSM8K $\rightarrow$ MATH $\rightarrow$ AIME.
This easy-to-hard ordering provides a fixed reference sequence for
budgeted evaluation.
We report the resulting performance as $\text{score}_{\text{easy}}$, which is further used to compute regret by comparing it with the score obtained under the original ordering.
Importantly, the original and easy variants share identical problem content and differ only in order.

\paragraph{Budgets.}
All test papers are evaluated under a fixed global token budget.
During training, the budget is set to 1024 tokens.
At evaluation time, we consider both a 1024-token setting and a more restrictive 512-token setting.
The latter regime strongly penalizes inefficient early reasoning and highlights the importance of selective skipping and early stopping.

\section{Baselines}
\label{app:baselines}

This appendix documents the baselines used in our experiments, with an emphasis on prompt-based controls under a strict \emph{global} token budget.
Unless otherwise stated, all baselines share the same evaluation protocol and decoding configuration; they differ only in the mechanism used to elicit budget awareness (e.g., whether the model explicitly plans, whether it must execute sequentially, and how skipping is triggered).
For reproducibility, we report the prompts verbatim.

\subsection{Simple and Explicit}
\label{app:baseline_simple}

This baseline provides a minimal, explicit instruction: solve the three problems sequentially under a strict global budget, and output \texttt{\textbackslash boxed\{\colorbox{gray!20}{NA}\}} when choosing to give up.
It evaluates the model's \emph{implicit} ability to trade off depth of reasoning against the opportunity cost of consuming budget on earlier questions, without any dedicated planning stage.

\begin{tcolorbox}[
    breakable,
    colback=gray!5, 
    colframe=gray!50!black, 
    title=\textbf{Simple and Explicit Prompt},
    fonttitle=\bfseries,
    arc=1mm
]
\footnotesize
The following text contains three mathematics problems. Please solve them sequentially.\\
\textbf{Constraint}: You have a strict limit of \textbf{1024 tokens}. Your goal is to \textbf{answer as many questions correctly as possible} within this limit.\\
For each question, please follow these steps to answer:

\begin{itemize}
    \item Provide step-by-step reasoning. You \textbf{MUST wrap your final answer in \textbackslash boxed\{\}} (e.g., \textbackslash boxed\{42\}).
    \item If you think the question is too difficult: \textbf{Do NOT provide reasoning}. Directly output \textbackslash boxed\{NA\}.
\end{itemize}
\textbf{Content}:
\begin{itemize}
    \item (Normal): [Reasoning] ... The final answer is \textbackslash boxed\{...\}.
    \item (Give up): \textbackslash boxed\{NA\}
\end{itemize}
- Ensure you strictly follow this format.

\end{tcolorbox}

\subsection{Plan-and-Solve}
\label{app:baseline_plan}

This baseline augments the above instruction with an explicit \emph{global meta-plan} that must be produced before any detailed reasoning.
The model is required to (i) scan all problems, (ii) estimate whether the budget suffices, and (iii) prioritize attempts based on a qualitative ROI notion.
In contrast to the simple prompt, the key difference is that budgeting decisions are made explicitly and upfront, rather than emerging implicitly during generation.

\begin{tcolorbox}[
    breakable,
    colback=gray!5, 
    colframe=gray!50!black, 
    title=\textbf{Plan-and-Solve Prompt},
    fonttitle=\bfseries,
    arc=1mm
]
\footnotesize
The following text contains three mathematics problems. You are taking a timed exam with a \textbf{strict global limit of 1024 tokens}. \\
\textbf{Your Goal}: \textbf{Maximize the total number of correct answers} within this budget. (Quality $>$ Quantity, but don't waste time on impossible tasks).

\vspace{0.5em}
\textbf{Strategy: Plan-First, Then Solve}\\
Before answering any specific question, you MUST perform a \textbf{Global Meta-Cognitive Assessment}:
\begin{enumerate}
    \item \textbf{Scan}: Briefly assess the difficulty of all three problems.
    \item \textbf{Budget}: Estimate if you have enough tokens to solve all of them. 1024 tokens is very tight!
    \item \textbf{Prioritize}: Decide which problems yield the highest ROI (Return on Investment). \textbf{Strategically Skip} hard/lengthy problems to save tokens for easier ones.
\end{enumerate}

\textbf{Response Format}:\\
\textbf{Step 1: Output your plan.}\\
\texttt{<Meta-Plan>}\\
Analysis: [Briefly assess P1, P2, P3 difficulty and estimated cost]\\
Strategy: [e.g., Attempt P1, Skip P2 (too long), Attempt P3]\\
\texttt{</Meta-Plan>}

\vspace{0.5em}
\textbf{Step 2: Solve sequentially based on your strategy.}\\
For each question:
\begin{itemize}
    \item If your strategy is \textbf{Attempt}: Provide concise step-by-step reasoning. You \textbf{MUST wrap your final answer in \textbackslash boxed\{\}}.
    \item If your strategy is \textbf{Skip}: Do \textbf{NOT} generate any reasoning. \textbf{Immediately output \textbackslash boxed\{NA\}} to conserve tokens.
\end{itemize}

\end{tcolorbox}

\subsection{Least-to-Most}
\label{app:baseline_least}

This baseline enforces \emph{strict sequential execution} while still allowing a limited look-ahead assessment.
The model must answer problems in the given order, but is encouraged to skip an early high-cost problem to preserve budget for later ones.
Relative to Plan-and-Solve, this baseline stresses the irreversibility of sequential allocation: even with global awareness, the execution order is fixed.

\begin{tcolorbox}[
    breakable,
    colback=gray!5, 
    colframe=gray!50!black, 
    title=\textbf{Least-to-Most Prompt},
    fonttitle=\bfseries,
    arc=1mm
]
\footnotesize
The following text contains three mathematics problems. You are taking a timed exam with a \textbf{strict global limit of 1024 tokens}. \\
\textbf{Your Goal}: \textbf{Maximize the total number of correct answers} within this budget.

\vspace{0.5em}
\textbf{Constraint: Strict Sequential Execution}\\
You \textbf{MUST} answer the problems strictly in the order they are presented: \textbf{Problem 1 $\to$ Problem 2 $\to$ Problem 3}. \\
You cannot change the order. However, you can (and should) strategically \textbf{SKIP} earlier problems if they are too costly, to save tokens for easier problems later.

\vspace{0.5em}
\textbf{Step 1: Global ROI Assessment (Look-Ahead Planning)}\\
Before answering Problem 1, analyze \textbf{ALL} three problems to form a plan:
\begin{enumerate}
    \item \textbf{Assess Cost/Reward}: Which problems are Easy (Low Cost), Medium, or Hard (High Cost)?
    \item \textbf{Knapsack Decision}: Given global tokens, if Problem 1 is very hard but Problem 2 and 3 are easy, the rational choice is to \textbf{SKIP Problem 1} (Output NA) to ensure you have budget for P2 and P3.
\end{enumerate}

\textbf{Step 2: Sequential Execution}\\
Execute your plan strictly in order (1, 2, 3). For each question:
\begin{itemize}
    \item \textbf{Check Plan \& Budget}: Based on your Step 1 plan and remaining tokens.
    \item \textbf{Action}:
    \begin{itemize}
        \item \textbf{Attempt}: Provide concise step-by-step reasoning. \textbf{MUST wrap final answer in \textbackslash boxed\{\}}.
        \item \textbf{Skip}: Do \textbf{NOT} provide reasoning. \textbf{Immediately output \textbackslash boxed\{NA\}} to conserve tokens.
    \end{itemize}
\end{itemize}

\end{tcolorbox}

\subsection{Plan-and-Budget with Jev}
\label{app:baseline_plan_budget}

This baseline adapts the separate planning and reasoning design of
Plan-and-Budget~\cite{lin2026plan} to our shared-budget, multi-problem
setting. We use Jev~\cite{jevmodels} as an external planner to predict
a cost/difficulty tag for each problem, and provide these tags to the
solver before generation. This is a training-free, level-tag
adaptation rather than a reproduction of the original method.

For each paper, Jev receives all three problems in their original
order, the shared output budget, and a description of the target
solver.

\paragraph{Structured Planner Input and Output.}
We query Jev through its structured classification API.
Each request contains a shared \texttt{state} and three
\texttt{questions}, one for each problem.
The state provides the target solver description, the shared
output-token budget, the task instruction, and all three problem
texts under \texttt{problems.problem\_1},
\texttt{problems.problem\_2}, and
\texttt{problems.problem\_3}.

Each question has \texttt{type="choice"} and contains two fields:
\texttt{instructions}, which specifies what to predict, and
\texttt{criteria}, which explicitly enumerates the candidate
labels and their meanings.
The keys of \texttt{criteria} are
\texttt{Level-0}, \texttt{Level-1}, \texttt{Level-2}, and
\texttt{Level-3}; their values are the descriptions shown below.
The same candidate set is supplied separately for each of the
three questions.

\begin{tcolorbox}[
    breakable,
    colback=gray!5,
    colframe=gray!50!black,
    title=\textbf{Jev: Task and Classification Instructions},
    fonttitle=\bfseries,
    arc=1mm
]
\footnotesize
\textbf{Task instruction in the shared state:}\\
Solve these three mathematics problems in the given order in one
response. Maximize the number of correct answers under the shared
hard output budget. Each answer must be boxed; an unsolvable problem
can be skipped with NA.

\medskip
\textbf{Instructions for question $i$, where $i\in\{1,2,3\}$:}\\
Assign a pre-reasoning cost/difficulty level to
\texttt{problems.problem\_}$i$ for the solver described in
\texttt{solver}, not for yourself. Estimate the tokens this solver
needs for a correct concise step-by-step solution, including its
final answer. Consider the shared budget and competing problems
when judging whether an unreliable attempt should be skipped.
Select one of the supplied levels; do not solve the problem.

\medskip
\textbf{Candidate labels and descriptions supplied in
\texttt{criteria}:}
\begin{itemize}
    \item \texttt{Level-0}: A correct step-by-step solution is
    expected to need fewer than 256 tokens.
    \item \texttt{Level-1}: A correct step-by-step solution is
    expected to need 256 to 512 tokens inclusive.
    \item \texttt{Level-2}: A correct step-by-step solution is
    expected to need more than 512 tokens.
    \item \texttt{Level-3}: Too difficult for the solver to answer
    reliably; skip reasoning and answer NA to preserve the shared
    budget. This means unlikely to solve, not merely long.
\end{itemize}
\end{tcolorbox}

Jev returns a structured JSON response containing one categorical
choice for each problem, rather than a reasoning trace or an XML
tag. The example below shows the returned choices with auxiliary
confidence, probability, and usage fields omitted.

\begin{tcolorbox}[
    breakable,
    colback=gray!5,
    colframe=gray!50!black,
    title=\textbf{Jev: Example Classification Output},
    fonttitle=\bfseries,
    arc=1mm
]
\footnotesize
\begin{verbatim}
{
  "model": "jev-1.13.0",
  "answers": {
    "problem_1": {
      "type": "choice",
      "choice": "Level-0"
    },
    "problem_2": {
      "type": "choice",
      "choice": "Level-1"
    },
    "problem_3": {
      "type": "choice",
      "choice": "Level-0"
    }
  }
}
\end{verbatim}
\end{tcolorbox}

\subsection{TALE-EP}
\label{app:baseline_tale}

This baseline adapts the estimate-then-solve principle of
TALE-EP~\cite{han2025token} to our multi-problem setting.
We use the solver itself, Qwen2.5-1.5B-Instruct, to predict one
cost/difficulty tag for each of the three problems before answering.
To match our evaluation interface, the estimation stage produces
Level-0/1/2/3 tags rather than numerical token-budget estimates.
This is a training-free level-tag adaptation of TALE-EP.

The baseline uses two calls to the same model.
The first call observes all three problems and predicts their tags
without solving them. The second call receives these tags alongside
the original problems and generates a single, ordered whole-paper
solution. Level-0/1/2 indicate expected solution lengths, while
Level-3 instructs the solver to skip the corresponding problem.
No per-problem hard decoding limits are imposed.

\paragraph{Estimation Prompt.}
In the following template, $B$ is replaced by the total evaluation
budget. The three problem texts are appended in their original
order, without reference answers or dataset-source labels.

\begin{tcolorbox}[
    breakable,
    colback=gray!5,
    colframe=gray!50!black,
    title=\textbf{TALE-EP: Estimation Prompt},
    fonttitle=\bfseries,
    arc=1mm
]
\footnotesize
Analyze the three problems below before solving them.
Predict one cost/difficulty level for each problem for yourself,
the solver. Estimate the tokens needed for a concise correct
solution including its final answer.

The three problems share a hard total output budget of $B$ tokens,
including this prediction step and the later whole-paper solution.
Each correct answer earns one point. Consider the competing problems
when deciding whether an unreliable attempt should be skipped.
\begin{itemize}
    \item Level-0: a correct solution is expected to require fewer
    than 256 output tokens.
    \item Level-1: a correct solution is expected to require
    256 to 512 output tokens, inclusive.
    \item Level-2: a correct solution is expected to require more
    than 512 output tokens.
    \item Level-3: too difficult for you to solve reliably;
    skip reasoning and answer NA. This is a solve-or-skip decision,
    not just a longer token range.
\end{itemize}
Do not solve the problems or explain your choices.
Output exactly three XML tags, one per line, corresponding to
Problems 1, 2, 3 in order. Use
\texttt{<predicted\_level>Level-X</predicted\_level>},
replacing X with 0, 1, 2, or 3.
Output only those three tags.

\medskip
\textbf{Problem 1}: [Problem 1 text]\\
\textbf{Problem 2}: [Problem 2 text]\\
\textbf{Problem 3}: [Problem 3 text]
\end{tcolorbox}

\subsection{ROI-Reasoning Prompt}
\label{app:roi_prompt}

Figure~\ref{fig:prompt_structure} shows the complete input organization
used by ROI-Reasoning. The instruction states the shared budget and
requires a predicted cost level before reasoning, while the three
problems are appended as a separate ordered block. For other
evaluation budgets, only the numerical budget value is replaced.

\begin{figure}[t]
\centering
\footnotesize
\begin{tcolorbox}[
  enhanced,
  title={\textbf{\large FULL INPUT PROMPT OVERVIEW}},
  colframe=acadblue,
  colback=white,
  colbacktitle=acadblue,
  fonttitle=\bfseries,
  boxrule=1.2pt,
  arc=2mm,
  boxsep=3pt,
  left=5pt,right=5pt,top=5pt,bottom=5pt
]
\begin{tcolorbox}[
  enhanced,
  title={\textbf{BLOCK 1: Instruction}},
  colframe=acadblue!80!white,
  colbacktitle=acadblue!10!white,
  coltitle=acadblue,
  colback=white,
  fonttitle=\bfseries,
  boxrule=0.8pt,
  arc=1mm,
  left=2pt,right=2pt,top=2pt,bottom=2pt,
  fontupper=\sffamily
]
\scriptsize
The following text contains three problems. Please solve them
sequentially.

\begin{tcolorbox}[
  colback=constraintred!8!white,
  colframe=constraintred,
  boxrule=0.5pt,
  left=2pt,right=2pt,top=1pt,bottom=1pt,
  arc=0mm
]
\textbf{Constraint:} You have a strict limit of
\textcolor{constraintred}{\textbf{1024 tokens}}. Your goal is to
\textbf{answer as many questions correctly as possible} within this
limit.
\end{tcolorbox}

\textbf{Step 1: Predict Length \& Difficulty}
\begin{itemize}[
  leftmargin=1.5em,topsep=0pt,partopsep=0pt,
  itemsep=0.05em,parsep=0.05em
]
  \item Before answering, estimate token length and difficulty for
  every problem.
  \item \textbf{Level-0}: $<256$ tokens.
  \item \textbf{Level-1}: $256$--$512$ tokens.
  \item \textbf{Level-2}: $>512$ tokens.
  \item \textbf{Level-3}: Too difficult (skip reasoning and answer
  \texttt{NA}).
\end{itemize}

\textbf{Step 2: Solve \& Box}
\begin{itemize}[
  leftmargin=1.5em,topsep=0pt,partopsep=0pt,
  itemsep=0.05em,parsep=0.05em
]
  \item If \textbf{Level 0--2}: provide step-by-step reasoning and
  wrap the final answer in \texttt{\textbackslash boxed\{\}}.
  \item If \textbf{Level 3}: do not provide reasoning; directly output
  \texttt{\textbackslash boxed\{NA\}}.
\end{itemize}

\begin{tcolorbox}[
  colback=acadblue!5!white,
  colframe=acadblue!40!white,
  boxrule=0.5pt,
  arc=0mm,
  left=2pt,right=2pt,top=1pt,bottom=1pt,
  title={\textbf{Strict Format Requirements:}},
  fonttitle=\bfseries\scriptsize,
  coltitle=black,
  colbacktitle=acadblue!15!white
]
\begin{itemize}[
  leftmargin=1.5em,topsep=0pt,partopsep=0pt,
  itemsep=0.05em,parsep=0.05em
]
  \item Start each problem with
  \texttt{<predicted\_level>Level-X</predicted\_level>}.
  \item Output either reasoning followed by
  \texttt{\textbackslash boxed\{\ldots\}}, or
  \texttt{\textbackslash boxed\{NA\}} when giving up.
\end{itemize}
\end{tcolorbox}
\end{tcolorbox}

\textcolor{gray}{\textbf{$\Downarrow$ (followed by)}}

\begin{tcolorbox}[
  enhanced,
  title={\textbf{BLOCK 2: Problems}},
  colframe=gray!80!white,
  colbacktitle=gray!15!white,
  coltitle=black,
  colback=acadgray!70!white,
  fonttitle=\bfseries,
  boxrule=0.8pt,
  arc=1mm,
  fontupper=\ttfamily\scriptsize
]
Problem 1: [Text content of Problem 1\ldots]\par
Problem 2: [Text content of Problem 2\ldots]\par
Problem 3: [Text content of Problem 3\ldots]
\end{tcolorbox}
\end{tcolorbox}
\caption{Structure of the ROI-Reasoning input prompt.}
\label{fig:prompt_structure}
\end{figure}

The two-block design exposes all problems before generation while
preserving the required sequential answer order. The predicted-level
tag makes the anticipated cost observable, and the explicit
\texttt{Level-3} branch provides a standardized zero-cost refusal
action.

\subsection{Greedy Knapsack}
\label{app:baseline_greedy}

\paragraph{Reference Baseline (Predict-then-Optimize).}
A conventional approach to resource-constrained multi-task reasoning follows the \emph{predict-then-optimize} paradigm.
Under this framework, the model first predicts, for each problem, its anticipated difficulty or computational cost, and then allocates a fixed budget across problems by solving a knapsack-style optimization problem.
This idea has been explored in prior work as a general strategy for computation allocation~\cite{brown2026predictiveschedulingefficientinferencetime}.

Following this paradigm, we include a \emph{Greedy Knapsack} baseline as a simple reference.
Rather than learning budget-aware generation end-to-end, this baseline decouples prediction and execution: it uses predicted difficulty signals to guide a post-hoc greedy selection of problems under a fixed token budget.
We treat this baseline as a lightweight heuristic for cost-based allocation.

Concretely, the model first generates candidate solutions for all problems in a test paper.
A greedy knapsack-style procedure is then applied to retain only a subset of problems whose estimated costs fit within the budget; rejected problems are replaced by \texttt{\textbackslash boxed\{NA\}}.
Unlike our method, this baseline does not regulate reasoning length during generation and relies on external selection after generation.

\paragraph{Cost Estimation.}
We interpret the discrete difficulty tags predicted by the MFT model as proxies for computational cost.
For greedy sorting, each predicted tag \texttt{Level-}\(k\) is mapped to a scalar expected token usage \(\hat{c}(k)\):
\begin{equation}
    \hat{c}(k) =
    \begin{cases}
    \frac{0+256}{2}=128 & \text{if } k=0, \\
    \frac{256+512}{2}=384 & \text{if } k=1, \\
    \frac{512+1024}{2}=768 & \text{if } k=2.
    \end{cases}
\end{equation}
These values correspond to the approximate centroids of the token ranges used in our difficulty categorization.
If a predicted tag is missing or malformed, we assign a default cost of \(\hat{c}=384\).

\newcommand{\PerProblemResultsTable}{%
\begin{table*}[t]
\centering
\setlength\tabcolsep{4pt}
\renewcommand{\arraystretch}{1.05}
\fontsize{9pt}{9pt}\selectfont

\resizebox{\textwidth}{!}{%
\begin{tabular}{llcccccccc}
\toprule
\multirow{2}{*}{\textbf{Method}} 
& \multirow{2}{*}{\textbf{Budget}}
& \multicolumn{4}{c}{\textbf{Medium}}
& \multicolumn{4}{c}{\textbf{Hard}} \\
\cmidrule(lr){3-6} \cmidrule(lr){7-10}
& 
& P1 & P2 & P3 & \textbf{Score}
& P1 & P2 & P3 & \textbf{Score} \\
\midrule

\textbf{DeepSeek-V3.2 (685B)} & 1024
& 0.60 & 0.45 & 0.28 & 1.33
& 0.32 & 0.29 & 0.26 & 0.87 \\
 & 512
& 0.48 & 0.29 & 0.11 & 0.88
& 0.19 & 0.17 & 0.13 & 0.49 \\

\textbf{GPT-4o-mini} & 1024
& 0.44 & 0.54 & 0.43 & 1.41
& 0.18 & 0.54 & 0.55 & 1.27 \\
 & 512
& 0.38 & 0.23 & 0.04 & 0.65
& 0.14 & 0.15 & 0.03 & 0.32 \\

\midrule

\textbf{Llama-3.1-8B-Instruct} & 1024
& 0.26 & 0.27 & 0.21 & 0.74
& 0.09 & 0.20 & 0.26 & 0.55 \\
 & 512
& 0.24 & 0.20 & 0.10 & 0.54
& 0.08 & 0.11 & 0.09 & 0.28 \\

\textbf{Qwen2.5-Math-7B-Instruct} & 1024
& 0.50 & 0.41 & 0.16 & 1.07
& 0.19 & 0.31 & 0.19 & 0.69 \\
 & 512
& 0.39 & 0.12 & 0.00 & 0.51
& 0.13 & 0.05 & 0.00 & 0.18 \\

\textbf{Qwen2.5-7B-Instruct} & 1024
& 0.45 & 0.50 & 0.32 & 1.27
& 0.17 & 0.49 & 0.41 & 1.07 \\
 & 512
& 0.39 & 0.22 & 0.03 & 0.64
& 0.14 & 0.12 & 0.03 & 0.29 \\

\textbf{Llama-3.2-3B-Instruct} & 1024
& 0.20 & 0.12 & 0.09 & 0.41
& 0.06 & 0.11 & 0.13 & 0.30 \\
 & 512
& 0.19 & 0.10 & 0.04 & 0.33
& 0.06 & 0.06 & 0.05 & 0.17 \\

\textbf{Qwen2.5-Math-1.5B-Instruct} & 1024
& 0.45 & 0.39 & 0.21 & 1.05
& 0.17 & 0.37 & 0.25 & 0.79 \\
 & 512
& 0.35 & 0.13 & 0.01 & 0.49
& 0.12 & 0.06 & 0.01 & 0.19 \\

\textbf{Qwen2.5-1.5B-Instruct} & 1024
& 0.24 & 0.21 & 0.15 & 0.60
& 0.07 & 0.22 & 0.20 & 0.49 \\
 & 512
& 0.21 & 0.10 & 0.02 & 0.33
& 0.07 & 0.06 & 0.02 & 0.15 \\

\midrule

\textbf{Qwen2.5-1.5B-Instruct} &  &  &  &  &  &  &  &  &  \\

\quad + Plan-and-Solve & 1024
& 0.20 & 0.12 & 0.08 & 0.40
& 0.07 & 0.10 & 0.11 & 0.28 \\
 & 512
& 0.11 & 0.03 & 0.01 & 0.15
& 0.03 & 0.01 & 0.01 & 0.05 \\

\quad + Least-to-Most & 1024
& 0.21 & 0.18 & 0.13 & 0.52
& 0.07 & 0.15 & 0.16 & 0.38 \\
 & 512
& 0.17 & 0.08 & 0.03 & 0.28
& 0.05 & 0.05 & 0.03 & 0.13 \\

\quad + Plan-and-Budget with Jev & 1024
& 0.21 & 0.18 & 0.12 & 0.51
& 0.07 & 0.17 & 0.16 & 0.41 \\
 & 512
& 0.20 & 0.09 & 0.02 & 0.31
& 0.07 & 0.06 & 0.02 & 0.16 \\

\quad + TALE-EP & 1024
& 0.16 & 0.15 & 0.11 & 0.43
& 0.06 & 0.15 & 0.16 & 0.38 \\
 & 512
& 0.15 & 0.10 & 0.03 & 0.28
& 0.06 & 0.08 & 0.05 & 0.18 \\

\quad + MFT & 1024
& 0.26 & 0.36 & 0.36 & 0.98
& 0.09 & 0.35 & 0.51 & 0.95 \\
 & 512
& 0.25 & 0.31 & 0.26 & 0.82
& 0.10 & 0.33 & 0.38 & 0.81 \\

\quad + MFT + Greedy Knapsack & 1024
& 0.26 & 0.36 & 0.35 & 0.97
& 0.09 & 0.35 & 0.52 & 0.96 \\
 & 512
& 0.25 & 0.31 & 0.27 & 0.83
& 0.08 & 0.34 & 0.41 & 0.83 \\

\rowcolor[RGB]{236,244,252}
\quad \textbf{+ ROI-REASONING} & 1024
& 0.29 & 0.43 & 0.41 & 1.13
& 0.10 & 0.41 & 0.61 & 1.12 \\
\rowcolor[RGB]{236,244,252}
 & 512
& 0.29 & 0.39 & 0.29 & 0.97
& 0.11 & 0.39 & 0.43 & 0.93 \\

\bottomrule
\end{tabular}}
\caption{Per-problem accuracy and overall score under different token budgets.
We report accuracy for each problem position (P1--P3) and the resulting total score on Medium and Hard test papers.}
\label{tab:acc_by_budget}
\end{table*}
}

\paragraph{Problem Extraction and Reordering.}
Because the model processes all three problems as a single sequence, we allow it to generate responses without a hard cutoff and then parse the raw output into per-problem segments.
We use the \texttt{\textbackslash boxed\{\}} delimiter to identify boundaries: the text segment ending at the \(i\)-th valid \texttt{\textbackslash boxed} closure is treated as the solution to the \(i\)-th problem.
This yields a set of candidate solutions that can be reordered based on their estimated costs.

\paragraph{Selection Algorithm.}
Let \(\mathcal{P}=\{p_1,\dots,p_N\}\) denote the problems in a test paper and let \(B\) be the global token budget (e.g., \(B=1024\)).
The greedy selection procedure is as follows:
\begin{enumerate}
    \item \textbf{Estimate.} For each problem \(p_i\), extract its predicted difficulty tag and assign an estimated cost \(\hat{c}_i\).
    \item \textbf{Sort.} Reorder the problems to obtain a permutation \(\pi\) such that
    \(\hat{c}_{\pi(1)} \le \hat{c}_{\pi(2)} \le \dots \le \hat{c}_{\pi(N)}\).
    \item \textbf{Select.} Iteratively include problems in sorted order, retaining \(p_{\pi(j)}\) if
    \(\sum_{k=1}^{j}\hat{c}_{\pi(k)} \le B\).
    \item \textbf{Fill.} For selected problems, the full generated solution is kept; for rejected problems, the output is replaced by \texttt{\textbackslash boxed\{NA\}}.
\end{enumerate}

\paragraph{Hard Constraint Enforcement.}
Finally, to ensure consistency with strict-budget evaluation, the reassembled output is tokenized.
If its total length exceeds the budget \(B\), a hard truncation is applied.
This penalizes cases where the actual generation length significantly exceeds the estimated cost \(\hat{c}(k)\), potentially resulting in incomplete solutions at the end of the sequence.

\section{Per-Problem Accuracy Analysis}
\label{app:per_problem_accuracy}

In this section, we provide a finer-grained analysis of model performance by examining \emph{per-problem accuracy}.
While the main experiments evaluate performance at the test-paper level under a global budget constraint, it is also useful to understand how accuracy varies across problem positions within a paper.

Specifically, for each test paper, we record whether the model correctly solves the problem at each position (P1, P2, and P3).
This positional analysis helps disentangle overall score gains from ordering- and budget-allocation effects.
In particular, it reveals whether the improvements come from better preserving performance on later problems under tight budgets, rather than from uniform gains across all positions.

Table~\ref{tab:acc_by_budget} reports per-problem accuracy on both Medium and Hard test papers under 1024-token and 512-token budgets.

\PerProblemResultsTable

The position-wise results show that the improvement is not obtained by
sacrificing the first problem. Relative to MFT, ROI-Reasoning preserves or
improves P1 and produces its clearest gains at P2 and P3, where
premature budget exhaustion is most damaging. On Hard-1024 papers, for
example, P2/P3 accuracy rises from 0.35/0.51 to 0.41/0.61, increasing
the total score from 0.95 to 1.12.

\section{Extensions to Other Domains}
\label{app:generalization}

While the primary experiments demonstrate the efficacy of ROI-Reasoning on mathematical reasoning, a natural question is whether this budget-aware allocation capability generalizes to other domains. To investigate this, we conduct additional experiments on two distinct types of tasks under a strict global token budget: Multiple-Choice Question Answering (MCQA) and Code Generation. Cross-domain experiments follow the main mathematical-reasoning setup in Appendix Section~\ref{app:experimental_setup}; only domain-specific changes are reported below. TALE-EP and Plan-and-Budget with Jev follow the same
procedures described in Appendix ~\ref{app:baseline_tale}
and~\ref{app:baseline_plan_budget}, with domain-specific problem text
and answer formats. In code generation, we additionally evaluate
Plan-and-Budget with Jev using Qwen2.5-Coder-1.5B-Instruct. RARL uses
the same optimization hyperparameters as in the main
mathematical-reasoning experiments
(Table~\ref{tab:rarl_hyperparameters}).

\newcommand{\MCQAResultsTable}{%
\begin{table*}[t]
\centering
\setlength\tabcolsep{3pt}
\renewcommand{\arraystretch}{1}
\fontsize{7pt}{8.9pt}\selectfont
\resizebox{\textwidth}{!}{%
\begin{tabular}{lcccccccccc}
\toprule
\multirow{3}{*}{\textbf{Method}}
& \multicolumn{2}{c}{\textbf{Score$_{\mathrm{easy}}$}}
& \multicolumn{4}{c}{\textbf{Medium}}
& \multicolumn{4}{c}{\textbf{Hard}} \\
\cmidrule(lr){2-3} \cmidrule(lr){4-7} \cmidrule(lr){8-11}
& \multirow{2}{*}{512}
& \multirow{2}{*}{256}
& \multicolumn{2}{c}{512}
& \multicolumn{2}{c}{256}
& \multicolumn{2}{c}{512}
& \multicolumn{2}{c}{256} \\
\cmidrule(lr){4-5} \cmidrule(lr){6-7}
\cmidrule(lr){8-9} \cmidrule(lr){10-11}
& & &
\textbf{Score}$\uparrow$ & \textbf{Reg.}$\downarrow$
& \textbf{Score}$\uparrow$ & \textbf{Reg.}$\downarrow$
& \textbf{Score}$\uparrow$ & \textbf{Reg.}$\downarrow$
& \textbf{Score}$\uparrow$ & \textbf{Reg.}$\downarrow$ \\
\midrule
\multicolumn{11}{c}{\textbf{Large-Scale Models}} \\
\textbf{DeepSeek-V3.2 (685B)}
& 2.04 & 1.80 & \textbf{1.78} & 0.13 & \textbf{1.41} & 0.22
& \underline{1.35} & 0.34 & 0.80 & 0.56 \\
\textbf{GPT-4o-mini}
& 1.55 & 1.31 & 1.30 & 0.16 & 0.94 & 0.28
& 0.93 & 0.40 & 0.51 & 0.61 \\
\midrule
\multicolumn{11}{c}{\textbf{Open-Source Models (1.5B--8B)}} \\
\textbf{Llama-3.1-8B-Instruct}
& 1.33 & 1.05 & 0.98 & 0.26 & 0.60 & 0.43
& 0.62 & 0.53 & 0.18 & 0.83 \\
\textbf{Qwen2.5-7B-Instruct}
& 1.34 & 1.14 & 1.02 & 0.24 & 0.67 & 0.41
& 0.64 & 0.52 & 0.18 & 0.84 \\
\textbf{Llama-3.2-3B-Instruct}
& 1.06 & 0.88 & 0.77 & 0.27 & 0.48 & 0.45
& 0.48 & 0.55 & 0.14 & 0.84 \\
\textbf{Qwen2.5-1.5B-Instruct}
& 0.78 & 0.63 & 0.48 & 0.38 & 0.31 & 0.51
& 0.29 & 0.63 & 0.06 & 0.90 \\
\quad + Plan-and-Solve
& 0.67 & 0.27 & 0.50 & 0.25 & 0.15 & 0.44
& 0.37 & 0.45 & 0.06 & 0.78 \\
\quad + Least-to-Most
& 0.66 & 0.59 & 0.48 & 0.27 & 0.37 & 0.37
& 0.43 & 0.35 & 0.23 & 0.61 \\
\quad + TALE-EP
& 1.11 & 1.10 & 1.03 & \underline{0.07} & 0.98 & 0.11
& 1.01 & \underline{0.09} & 0.93 & 0.15 \\
\quad + Plan-and-Budget with Jev
& 1.13 & 1.12 & 1.04 & 0.08 & 1.02 & \underline{0.09}
& 1.00 & 0.12 & \underline{0.97} & \underline{0.13} \\
\rowcolor[RGB]{236,244,252}
\quad\textbf{ + ROI-Reasoning}
& 1.37 & 1.36
& \underline{1.39} & \textbf{-0.01}
& \underline{1.38} & \textbf{-0.01}
& \textbf{1.39} & \textbf{-0.01}
& \textbf{1.38} & \textbf{-0.01} \\
\bottomrule
\end{tabular}}
\caption{Performance comparison on Multiple-Choice Question Answering under different difficulty levels and token budgets. We report Score (maximum $=3$), Score$_{\mathrm{easy}}$, and Reference-Order Regret on Medium and Hard test papers, each consisting of three questions, under 512-token and 256-token constraints. The best and second-best values in each Score and Regret column are highlighted in \textbf{bold} and \underline{underlined}, respectively. }
\label{tab:mcq_results}
\end{table*}%
}
\newcommand{\CodeResultsTable}{%
\begin{table*}[t]
\centering
\setlength\tabcolsep{3pt}
\renewcommand{\arraystretch}{1.05}
\fontsize{7pt}{8.9pt}\selectfont
\resizebox{\textwidth}{!}{%
\begin{tabular}{lcccccccccc}
\toprule
\multirow{3}{*}{\textbf{Method}}
& \multicolumn{2}{c}{\textbf{Score$_{\mathrm{easy}}$}}
& \multicolumn{4}{c}{\textbf{Medium}}
& \multicolumn{4}{c}{\textbf{Hard}} \\
\cmidrule(lr){2-3} \cmidrule(lr){4-7} \cmidrule(lr){8-11}
& \multirow{2}{*}{512}
& \multirow{2}{*}{256}
& \multicolumn{2}{c}{512}
& \multicolumn{2}{c}{256}
& \multicolumn{2}{c}{512}
& \multicolumn{2}{c}{256} \\
\cmidrule(lr){4-5} \cmidrule(lr){6-7}
\cmidrule(lr){8-9} \cmidrule(lr){10-11}
& & &
\textbf{Score}$\uparrow$ & \textbf{Reg.}$\downarrow$
& \textbf{Score}$\uparrow$ & \textbf{Reg.}$\downarrow$
& \textbf{Score}$\uparrow$ & \textbf{Reg.}$\downarrow$
& \textbf{Score}$\uparrow$ & \textbf{Reg.}$\downarrow$ \\
\midrule
\multicolumn{11}{c}{\textbf{Large-Scale Models}} \\
\textbf{DeepSeek-V3.2 (685B)}
& 2.03 & 1.81 & \textbf{1.96} & \underline{0.03}
& \textbf{1.64} & \textbf{0.09}
& \textbf{1.85} & 0.09
& \textbf{1.57} & \underline{0.13} \\
\textbf{GPT-4o-mini}
& 1.61 & 1.21 & \underline{1.59} & \textbf{0.01}
& 1.10 & \textbf{0.09}
& \underline{1.77} & \textbf{-0.10}
& \underline{1.45} & \textbf{-0.20} \\
\midrule
\multicolumn{11}{c}{\textbf{Open-Source Models (1.5B--8B)}} \\
\textbf{Llama-3.1-8B-Instruct}
& 1.41 & 1.23 & 1.12 & 0.21 & 0.98 & 0.20
& 0.98 & 0.30 & 0.74 & 0.40 \\
\textbf{Qwen2.5-7B-Instruct}
& 1.61 & 1.39 & 1.40 & 0.13 & \underline{1.15} & 0.17
& 1.13 & 0.30 & 0.77 & 0.45 \\
\textbf{Llama-3.2-3B-Instruct}
& 0.09 & 0.03 & 0.04 & 0.56 & 0.02 & 0.33
& 0.01 & 0.89 & 0.01 & 0.67 \\
\textbf{Qwen2.5-Coder-1.5B-Instruct}
& 1.37 & 1.32 & 1.10 & 0.20 & 0.88 & 0.33
& 0.43 & 0.69 & 0.09 & 0.93 \\
\quad + Plan-and-Budget with Jev
& 1.36 & 1.24 & 1.26 & 0.07 & 0.97 & 0.22
& 0.73 & 0.46 & 0.11 & 0.91 \\
\textbf{Qwen2.5-1.5B-Instruct}
& 0.79 & 0.68 & 0.50 & 0.37 & 0.42 & 0.38
& 0.11 & 0.86 & 0.02 & 0.97 \\
\quad + Plan-and-Solve
& 1.01 & 0.95 & 0.69 & 0.32 & 0.58 & 0.39
& 0.37 & 0.63 & 0.33 & 0.65 \\
\quad + Least-to-Most
& 0.92 & 0.82 & 0.58 & 0.37 & 0.55 & 0.33
& 0.44 & 0.52 & 0.30 & 0.63 \\
\quad + TALE-EP
& 0.68 & 0.53 & 0.50 & 0.26 & 0.46 & 0.13
& 0.18 & 0.74 & 0.14 & 0.74 \\
\quad + Plan-and-Budget with Jev
& 0.69 & 0.55 & 0.55 & 0.20 & 0.46 & 0.16
& 0.20 & 0.71 & 0.18 & 0.67 \\
\rowcolor[RGB]{236,244,252}
\quad\textbf{ + ROI-Reasoning}
& 1.35 & 0.87 & 1.31 & \underline{0.03}
& 0.76 & 0.13
& 1.34 & \underline{0.01}
& 0.73 & 0.16 \\
\bottomrule
\end{tabular}}
\caption{Performance comparison on Code Generation tasks under different difficulty levels and token budgets. We report Score (Pass@1, maximum $=3$), Score$_{\mathrm{easy}}$, and Reference-Order Regret on Medium and Hard test papers under 512-token and 256-token constraints. The best and second-best values in each Score and Regret column are highlighted in \textbf{bold} and \underline{underlined}, respectively. }
\label{tab:code_results}
\end{table*}%
}

\subsection{Multiple-Choice Question Answering}
\label{app:mcq}

To evaluate the cross-domain adaptation of ROI-Reasoning, we extend our evaluation to Multiple-Choice Question Answering (MCQA). Unlike mathematical word problems, MCQA tasks typically involve shorter reasoning chains but require broader factual knowledge and the ability to distinguish between distracting options under strict resource constraints.

\paragraph{Data Construction and Splits.} 
We curate a multi-tier difficulty benchmark using three representative datasets:
\begin{itemize}
    \item \textbf{Easy: ARC (AI2 Reasoning Challenge) \cite{clark2018thinksolvedquestionanswering}.} We utilize the ARC set, which focuses on elementary and middle-school level science questions requiring basic reasoning.
    \item \textbf{Medium: MMLU-PRO \cite{wang2024mmlu}.} As a more robust and challenging evolution of the original MMLU, MMLU-PRO is used for medium-difficulty tasks. We implement a \textbf{9:1 split} for training and testing to ensure the model adapts to the specific formatting and reasoning requirements of professional knowledge domains.
    \item \textbf{Hard: GPQA (Graduate-Level Google-Proof Q\&A) \cite{rein2023gpqa}.} For the hardest tier, we use the GPQA dataset, which consists of extremely challenging science questions written by experts. We strictly use the \textbf{GPQA-Diamond} set as our evaluation benchmark, while the non-Diamond portions of \textbf{GPQA-Extend} are utilized for the training phase.
\end{itemize}
In alignment with the characteristics of MCQA tasks, we evaluate all models under even more restrictive global budgets of \textbf{512} and \textbf{256} tokens to emphasize the importance of concise reasoning and strategic allocation. Consistent with our primary experiments, we evaluate the models on 5,000 synthesized test papers. This large-scale assessment ensures statistical robustness and provides a reliable measure of the model's meta-cognitive allocation across varying problem orderings in the MCQA domain.

\paragraph{Experimental Details and Baselines.}
To ensure a rigorous evaluation, we maintain consistency with the baselines used in our primary study, including both large-scale proprietary models and open-source counterparts. Within this comparative framework, we implement a strict scoring policy for these generalization tasks: any skipped questions (explicitly outputting \texttt{NA}) or incomplete responses due to budget truncation are scored as $0$. Furthermore, we deliberately do not train the model to perform random guessing, ensuring that the results reflect authentic reasoning and meta-cognitive allocation capabilities. 

Beyond these evaluation protocols, we employ a streamlined, two-step variant of the MFT procedure described in Appendix \ref{app:meta-cognitive} for these specific domains:
\begin{itemize}
    \item \textbf{Step 1: Formatting and Tag Alignment.} We first perform single-problem adaptation to ensure the model adheres to the required output syntax and accurately predicts difficulty tags for individual items.
    \item \textbf{Step 2: Sequence Synthesis and Refusal Learning.} To construct high-quality multi-problem training data, we decompose three-problem sequences into individual queries during sampling. We use the tag-aware model from Step 1 to generate high-quality independent solutions, which are then concatenated into full sequences. Concurrently, we implement \textit{Refusal Learning} by identifying individual problems with persistently low global accuracy and replacing their target outputs with a \texttt{Level-3} (skip) tag.
\end{itemize}
This simplified approach maintains the core benefits of meta-cognitive alignment while significantly enhancing data construction efficiency. For the final training, the models undergo full-parameter fine-tuning on a curated dataset comprising tens of thousands of synthesized sequences. The training is conducted for 3 epochs using a learning rate of $5 \times 10^{-5}$.

\MCQAResultsTable

\paragraph{Analysis of Results.}
Table~\ref{tab:mcq_results} shows that ROI-Reasoning achieves a higher Score than the best same-backbone, training-free baseline shown in the table across all four settings, with gains of 0.35--0.41 points (approximately 34--42\%). It also attains the lowest reference-order Regret in every setting, improving on the best same-backbone baseline by 0.08--0.14. The largest Score gain occurs on Hard-256 papers: compared with Plan-and-Budget with Jev, Score increases from 0.97 to 1.38, while Regret falls from 0.13 to $-0.01$. These results show that the learned allocation policy remains effective across both budgets and paper orders in MCQA.

\paragraph{Understanding Negative Reference-Order Regret.}
Table~\ref{tab:mcq_results} reports a slightly negative Regret of $-0.01$ for the full ROI-Reasoning model across the MCQA settings. The MFT-only variant is omitted from this method-comparison table, but exhibits more pronounced negative values, reaching $-0.23$. A negative Regret means that the original paper order yields a higher Score than the easy-to-hard reference order.

\begin{table*}[t]
\centering
\setlength\tabcolsep{3pt}
\renewcommand{\arraystretch}{1.05}
\fontsize{8pt}{10pt}\selectfont
\begin{tabular}{lcccccc}
\toprule
\multirow{2}[2]{*}{\textbf{Paper Type}} & \multicolumn{3}{c}{\textbf{Accuracy(\%)}} & \multicolumn{3}{c}{\textbf{Average Tokens}} \\
\cmidrule(lr){2-4} \cmidrule(lr){5-7}
& \textbf{Easy} & \textbf{Medium} & \textbf{Hard} & \textbf{Easy} & \textbf{Medium} & \textbf{Hard} \\
\midrule
Easy Paper & 38.54 & 27.34 & 14.36 & 49.2 & 46.0 & 44.6 \\
Medium Paper     & 53.82 & 38.35 & 10.28 & 56.1 & 54.4 & 40.4 \\
Hard Paper & \textbf{61.55} & \textbf{46.63} & 9.94 & \textbf{58.6} & \textbf{58.6} & 37.9 \\
\bottomrule
\end{tabular}
\caption{Behavioral Analysis of MFT across different test paper orderings. By identifying and strategically curtailing effort on Hard questions early, the model reallocates saved tokens to dramatically boost the accuracy of Easy and Medium questions.}
\label{tab:negative_regret_analysis}
\end{table*}
As analyzed in Table~\ref{tab:negative_regret_analysis}, MFT exhibits a budget-reallocation pattern when a Hard problem appears first. Its learned cost assessment leads it to curtail reasoning on this low-ROI problem, spending only 37.9 tokens on average. Conceptually, the model follows the resource-management strategy of an experienced test-taker: rather than exhausting its budget on a question with a low chance of success, it accepts a local loss and preserves tokens for the remaining problems.

This selective curtailment leaves more room for subsequent Medium and Easy questions, on which MFT spends an average of 58.6 tokens each. Compared with easy-first papers, accuracy on Easy questions rises from 38.54\% to 61.55\%, and accuracy on Medium questions rises from 27.34\% to 46.63\%. Meanwhile, Hard-question accuracy falls from 14.36\% to 9.94\%. Thus, a 4.42-percentage-point loss on Hard questions accompanies gains of 23.01 and 19.29 percentage points on Easy and Medium questions, respectively. This trade-off helps explain how MFT can achieve a higher total Score under a hard-first order than under the easy-to-hard reference.

\subsection{Code Generation}
\label{app:code_generation}

To further validate the robustness of ROI-Reasoning, we evaluate its performance on code generation, a domain where structural integrity and logical completeness are paramount. Under strict token constraints, code tasks present an ``all-or-nothing" challenge: a truncated code block due to budget exhaustion is non-executable, resulting in a zero score regardless of its partial correctness.

\paragraph{Data Construction and Splits.} 
Following our multi-tier difficulty framework, we curate a comprehensive coding benchmark by integrating three representative datasets, each serving a specific difficulty level:
\begin{itemize}
    \item \textbf{Easy: MBPP \cite{austin2021program}.} We utilize the sanitized version of MBPP, focusing on entry-level Python tasks. The prompts are simplified to core task descriptions with a single usage example to minimize initial token overhead.
    \item \textbf{Medium: HumanEval \cite{chen2021evaluating}.} This dataset consists of standard algorithmic problems. We present these tasks via a function signature and docstring format, requiring the model to synthesize the full implementation.
    \item \textbf{Hard: BigCodeBench \cite{zhuo2024bigcodebench}.} For the high-difficulty tier, we leverage BigCodeBench, which involves complex, library-heavy programming tasks. 
\end{itemize}

\newcommand{\HeterogeneousResultsTable}{%
\begin{table*}[t]
\centering
\setlength\tabcolsep{3pt}
\renewcommand{\arraystretch}{1}
\fontsize{7pt}{8.9pt}\selectfont

\resizebox{\textwidth}{!}{%
\begin{tabular}{lcccccccccc}
\toprule
\multirow{3}{*}{\textbf{Method}}
& \multicolumn{2}{c}{\textbf{Score$_{\mathrm{easy}}$}}
& \multicolumn{4}{c}{\textbf{Medium}}
& \multicolumn{4}{c}{\textbf{Hard}} \\
\cmidrule(lr){2-3} \cmidrule(lr){4-7} \cmidrule(lr){8-11}
& \multirow{2}{*}{1024}
& \multirow{2}{*}{512}
& \multicolumn{2}{c}{1024}
& \multicolumn{2}{c}{512}
& \multicolumn{2}{c}{1024}
& \multicolumn{2}{c}{512} \\
\cmidrule(lr){4-5} \cmidrule(lr){6-7}
\cmidrule(lr){8-9} \cmidrule(lr){10-11}
& & &
\textbf{Score}$\uparrow$ & \textbf{Reg.}$\downarrow$
& \textbf{Score}$\uparrow$ & \textbf{Reg.}$\downarrow$
& \textbf{Score}$\uparrow$ & \textbf{Reg.}$\downarrow$
& \textbf{Score}$\uparrow$ & \textbf{Reg.}$\downarrow$ \\
\midrule

\multicolumn{11}{c}{\textbf{Large-Scale Models}} \\

\textbf{DeepSeek-V3.2 (685B)}
& 51.35 & 39.77
& \textbf{39.47} & 0.23
& \textbf{25.49} & 0.36
& 27.10 & 0.47
& 14.94 & 0.62 \\

\textbf{GPT-4o-mini}
& 40.48 & 28.96
& \underline{37.77} & 0.07
& 18.59 & 0.36
& \textbf{35.04} & 0.13
& 10.43 & 0.64 \\

\midrule
\multicolumn{11}{c}{\textbf{Open-Source Models (1.5B--8B)}} \\

\textbf{Llama-3.1-8B-Instruct}
& 25.34 & 23.07
& 19.30 & 0.24
& 14.57 & 0.37
& 14.15 & 0.44
& 7.75 & 0.66 \\

\textbf{Qwen2.5-Math-7B-Instruct}
& 42.25 & 25.53
& 30.34 & 0.28
& 15.35 & 0.40
& 20.39 & 0.52
& 5.89 & 0.77 \\

\textbf{Qwen2.5-7B-Instruct}
& 41.21 & 31.00
& 34.89 & 0.15
& 18.38 & 0.41
& \underline{30.09} & 0.27
& 9.08 & 0.71 \\

\textbf{Llama-3.2-3B-Instruct}
& 15.87 & 14.76
& 11.48 & 0.28
& 9.46 & 0.36
& 8.22 & 0.48
& 5.02 & 0.66 \\

\textbf{Qwen2.5-Math-1.5B-Instruct}
& 37.79 & 24.43
& 29.50 & 0.22
& 14.70 & 0.40
& 22.89 & 0.39
& 6.06 & 0.75 \\

\textbf{Qwen2.5-1.5B-Instruct}
& 18.74 & 13.76
& 16.58 & 0.12
& 9.63 & 0.30
& 13.36 & 0.29
& 4.73 & 0.66 \\

\quad + Plan-and-Solve
& 15.04 & 7.83
& 11.37 & 0.24
& 4.57 & 0.42
& 7.60 & 0.49
& 1.70 & 0.78 \\

\quad + Least-to-Most
& 16.83 & 11.08
& 14.09 & 0.16
& 8.16 & 0.26
& 10.31 & 0.39
& 3.89 & 0.65 \\

\quad + MFT
& 25.66 & 22.13
& 25.46 & \textbf{0.01}
& 21.48 & \underline{0.03}
& 24.74 & 0.04
& 21.78 & \underline{0.02} \\

\quad + MFT + Greedy Knapsack
& 25.72 & 21.93
& 25.24 & \underline{0.02}
& 21.72 & \textbf{0.01}
& 24.93 & \underline{0.03}
& \underline{21.91} & \textbf{0.00} \\

\rowcolor[RGB]{236,244,252}
\quad\textbf{ + ROI-Reasoning}
& 29.62 & 27.50
& 29.46 & \textbf{0.01}
& \underline{25.45} & 0.07
& 29.26 & \textbf{0.01}
& \textbf{25.34} & 0.08 \\

\bottomrule
\end{tabular}}

\caption{Performance under heterogeneous score on mathematical reasoning. AIME, MATH, and GSM8K problems are assigned 3, 2, and 1 points, respectively, and scores are normalized to a 100-point scale. We report Score, Score$_{\text{easy}}$, and Reference-Order Regret on Medium and Hard test papers under 1024-token and 512-token budgets. Best and second-best Score and Regret values are highlighted with
\textbf{bold} and \underline{underlining}.}
\label{tab:heterogeneous_reward_results}
\end{table*}
}

To ensure all tasks are executable in a standard headless server environment for fair evaluation, we implement a strict filtering pipeline for BigCodeBench. We exclude problems that involve: (i) \textit{Graphical Visualization} (e.g., Matplotlib, Seaborn); (ii) \textit{External Network Dependencies} (e.g., Requests, Socket); (iii) \textit{Restricted System Commands} (e.g., Subprocess, os.system); and (iv) \textit{Complex File I/O} (e.g., \texttt{pd.read\_csv}) which depends on external local data assets that cannot be consistently provided during automated evaluation.

The final consolidated dataset is shuffled and partitioned into a 9:1 split for training and testing. This process yields a robust benchmark where each test paper consists of three problems, evaluated under global budgets of 512 and 256 tokens. Considering the substantial computational overhead of execution-based evaluation, we evaluate on 200 unique test papers. Although the number of papers is smaller than in the MCQA domain, this collection involves 600 problem instances (3 per paper) sampled with replacement from our test set. This ensemble ensures that every problem in the test set is evaluated multiple times in different contextual orderings, providing a comprehensive and representative assessment of the model's budget-aware performance.

\paragraph{Experimental Details and Baselines.}
The experimental setup and baselines remain consistent with previous sections. We employ the same two-step MFT variant—comprising \textit{Tag Alignment} and \textit{Refusal Learning}—to instill budget-aware decision-making. However, unlike the MCQA domain and mathematical reasoning domains, the curated coding dataset is relatively small (on the scale of hundreds of samples). To prevent catastrophic forgetting of the model's underlying programming logic, we shift from full-parameter tuning to LoRA fine-tuning for this task. Specifically, we use a learning rate of $2 \times 10^{-4}$ and a rank of $r=8$, training for 3 epochs. This approach ensures the model adapts to our meta-cognitive framework while preserving its fundamental code synthesis capabilities.

\CodeResultsTable

\paragraph{Analysis of Results.}
Table~\ref{tab:code_results} shows that ROI-Reasoning exceeds the best same-backbone, training-free baseline in Score by 0.18--0.90 points across the four settings. Its reference-order Regret is lower in three settings and ties TALE-EP on Medium-256 papers. Under the tight Hard-256 setting, ROI-Reasoning scores 0.73, compared with 0.33 for Plan-and-Solve, the strongest same-backbone baseline by Score; its Regret is 0.16, compared with 0.63 for Least-to-Most, the strongest by Regret.

We also compare against a code-specialized backbone paired with an external planner: Qwen2.5-Coder-1.5B-Instruct with Plan-and-Budget with Jev. Despite using a general-purpose backbone, ROI-Reasoning achieves a higher Score in three of the four settings and a lower Regret in all four. The difference is most pronounced on Hard-256 papers, where Score rises from 0.11 to 0.73 and Regret falls from 0.91 to 0.16. The specialized baseline retains a higher Score on Medium-256 papers (0.97 versus 0.76), so the advantage is especially clear when difficult code problems appear first.

\section{Heterogeneous Evaluation Scoring}
\label{app:heterogeneous-reward}
This appendix reports a post-hoc robustness check under an alternative scoring scheme. We keep the trained models and generated outputs fixed, and only re-score the same mathematical test papers: AIME = 3, MATH = 2, and GSM8K = 1. 
For a test paper $x_{1:N}$, let $w_i$ denote the source-specific weight of problem $x_i$. 
We define the weighted normalized score as
\[
S^{\mathrm{wt}}_B(\pi; x_{1:N})
=
100 \cdot
\frac{\sum_{i=1}^N w_i \mathbf{1}\{\mathrm{correct}_i\}}
{\sum_{i=1}^N w_i}.
\]

Following the main text, we report the same reference-order regret using weighted scores:
\[
\mathrm{Reg}^{\mathrm{wt}}_B(\pi)
=
\frac{S^{\mathrm{wt,easy}}_B(\pi)-S^{\mathrm{wt}}_B(\pi)}
{S^{\mathrm{wt,easy}}_B(\pi)}.
\]

\HeterogeneousResultsTable

Table~\ref{tab:heterogeneous_reward_results} preserves the main ranking trend. Under strict budgets, MFT-based variants achieve higher weighted Score and lower reference-order Regret than structured reasoning baselines. ROI-Reasoning consistently improves over MFT and attains the best weighted Score in the most constrained Hard-512 setting. Although large frontier models often obtain relatively high Score$_{\text{easy}}$, their performance drops more sharply under the original order, resulting in larger reference-order Regret. In contrast, MFT and ROI-Reasoning remain more stable across settings.

\section{Case Studies}
\label{app:case_studies}
To complement the aggregate results in the main text, we provide three qualitative case studies that illustrate how ROI-Reasoning behaves under a shared token budget. 
These examples are intended to shed light on the mechanisms underlying the quantitative gains, rather than to serve as additional benchmark evaluations. 
Specifically, the first case highlights successful meta-cognitive resource allocation, the second shows that the learned behavior goes beyond binary solve-or-skip decisions and extends to adaptive shortening of still-correct reasoning, and the third presents a representative failure mode caused by cost underestimation. 
Taken together, these cases provide a more concrete view of both the strengths and limitations of ROI-Reasoning under budgeted inference.

\subsection{Meta-cognitive Resource Allocation}
\label{app:case_study_full}

This case study illustrates ROI-Reasoning's ability to perform intrinsic difficulty estimation and strategic decision-making. Faced with a complex geometry problem (Problem 1) and two simpler tasks (Problem 2 \& 3), the model opts to skip the former to stay within the strict 512-token budget.

\begin{tcolorbox}[
    breakable,
    colback=gray!5, 
    colframe=gray!50!black, 
    title=\textbf{Full Input Prompt (Constraint: 512 Tokens)},
    fonttitle=\bfseries,
    arc=1mm
]
\footnotesize
The following text contains three mathematics problems. Please solve them sequentially.\\
\textbf{Constraint}: You have a strict limit of \textbf{512 tokens}. Your goal is to \textbf{answer as many questions correctly as possible} within this limit.\\
For each question, please follow these steps to answer:

\textbf{Step 1: Predict Length \& Difficulty}\\
Before answering, estimate the token length and difficulty for every single problem:
\begin{enumerate}
    \item \textbf{Level-0}: $<$ 256 tokens
    \item \textbf{Level-1}: 256 - 512 tokens
    \item \textbf{Level-2}: $>$ 512 tokens
    \item \textbf{Level-3}: Too difficult (Skip reasoning, Answer NA)
\end{enumerate}

\textbf{Step 2: Solve \& Box}\\
- If \textbf{Level 0-2}: Provide step-by-step reasoning. You \textbf{MUST wrap your final answer in \textbackslash boxed\{\}} (e.g., \textbackslash boxed\{42\}).\\
- If \textbf{Level 3}: \textbf{Do NOT provide reasoning}. Directly output \textbackslash boxed\{NA\}.

\textbf{Strict Format Requirements}:\\
- Start with: \\
\texttt{<predicted\_level>Level-X</predicted\_level>} for each problem\\
- Content:
\begin{itemize}
    \item (Normal): [Reasoning] ... The final answer is \textbackslash boxed\{...\}.
    \item (Give up): \textbackslash boxed\{NA\}
\end{itemize}
- Ensure you strictly follow this XML format.

\noindent\rule{\textwidth}{0.4pt}

\textbf{Problem 1}: In $\triangle ABC$ points $D$ and $E$ lie on $\overline{AB}$ so that $AD < AE < AB$, while points $F$ and $G$ lie on $\overline{AC}$ so that $AF < AG < AC$. Suppose $AD = 4$, $DE = 16$, $EB = 8$, $AF = 13$, $FG = 52$, and $GC = 26$. Let $M$ be the reflection of $D$ through $F$, and let $N$ be the reflection of $G$ through $E$. The area of quadrilateral $DEGF$ is $288$. Find the area of heptagon $AFNBCEM$.

\textbf{Problem 2}: Compute $\arcsin \left( -\frac{1}{2} \right).$ Express your answer in radians.

\textbf{Problem 3}: The rainstorm flooded the wetlands and washed Phineas Frog 200 yards away from his home in the swamp. To return home, he will need to hop and swim his way back. If he hops on land at a speed of 20 yards per minute, and swims through water at a speed of 10 yards per minute, how long will it take Phineas, in minutes, to return home if half of the distance is on land and the other half is in water?
\end{tcolorbox}

\begin{tcolorbox}[
    breakable,
    colback=blue!5, 
    colframe=blue!50!black, 
    title=\textbf{Model Output},
    fonttitle=\bfseries,
    arc=1mm
]
\small
\textbf{Solution to Problem 1:} \\
\texttt{<predicted\_level>Level-3</predicted\_level>} \\
\fbox{NA}

\textbf{Solution to Problem 2:} \\
\texttt{<predicted\_level>Level-0</predicted\_level>} \\
We need to compute $\arcsin \left( -\frac{1}{2} \right)$. The value of $\arcsin \left( -\frac{1}{2} \right)$ is $-\frac{\pi}{6}$ radians.\\
The answer is \fbox{$-\frac{\pi}{6}$}.

\textbf{Solution to Problem 3:} \\
\texttt{<predicted\_level>Level-0</predicted\_level>} \\
Phineas needs to hop and swim back to his home. The distance to be covered is 200 yards. Half of the distance is 100 yards on land and half is 100 yards in water.\\
The time to hop back on land is $100 \div 20 = 5$ minutes.\\
The time to swim back in water is $100 \div 10 = 10$ minutes.\\
The total time is $5 + 10 = 15$ minutes.\\
The answer is \fbox{15}.
\end{tcolorbox}

\begin{tcolorbox}[
    colback=green!5, 
    colframe=green!40!black, 
    title=\textbf{Ground Truth},
    fonttitle=\bfseries,
    arc=1mm
]
\small
\textbf{Problem 1:} \fbox{588} \quad \textbf{Problem 2:} \fbox{$-\pi/6$} \\ \textbf{Problem 3:} \fbox{15}
\end{tcolorbox}

\paragraph{Observation.} 
The model demonstrates an effective \textit{predict-then-reason} behavior. Problem 1 is an AIME-level geometry problem that typically requires extensive coordinate geometry or area-ratio calculations, likely exceeding the remaining token budget. By correctly assigning \texttt{Level-3}, the model bypasses the complex reasoning and successfully solves the subsequent two problems within the 512-token limit, achieving a total score of 2.

\subsection{Adaptive Reasoning Beyond Binary Decisions}
\label{app:adaptive_reasoning_case}

The following Hard-512 case illustrates that the gain of RARL is not driven by an early Level-3 skip. Both MFT and ROI-Reasoning attempt the first two problems and answer them correctly. The difference arises within attempted problems: ROI-Reasoning uses shorter, still-correct reasoning on the first two questions, which preserves enough budget to complete the third one.

\paragraph{Execution summary.}
\begin{itemize}
    \item \textbf{MFT-only execution:}
    \begin{itemize}
        \item Q1: Level-1, answer correct.
        \item Q2: Level-0, answer correct.
        \item Q3: output truncated by the global token limit before a complete boxed answer.
    \end{itemize}

    \item \textbf{ROI-Reasoning execution (ours):}
    \begin{itemize}
        \item Q1: Level-0, answer correct.
        \item Q2: Level-0, answer correct.
        \item Q3: Level-0, answer correct within the same global budget.
    \end{itemize}
\end{itemize}

Using the same local tokenizer, the first two solutions contain 297 estimated output tokens for MFT and 183 for ROI-Reasoning.

\begin{tcolorbox}[
    breakable,
    colback=gray!5,
    colframe=gray!50!black,
    title=\textbf{Input Problems (Hard-512)},
    fonttitle=\bfseries,
    arc=1mm
]
\small
\textbf{Problem 1.}
The amount of money a certain company makes can be represented by the graph $y=x^2-8x-33$, where $x$ is the quantity produced. What is the smallest quantity at which the company can produce to break even or make a profit?

\textbf{Problem 2.}
If $f(x)=x+2$ and $g(x)=x^2$, then for what value of $x$ does $f(g(x))=g(f(x))$? Express your answer as a common fraction.

\textbf{Problem 3.}
There are twice as many boys as girls at Dr. Wertz's school. If there are 60 girls and 5 students to every teacher, how many teachers are there?
\end{tcolorbox}

\begin{tcolorbox}[
    breakable,
    colback=blue!5,
    colframe=blue!50!black,
    title=\textbf{MFT Output},
    fonttitle=\bfseries,
    arc=1mm
]
\footnotesize
\textbf{Solution to Problem 1:}\\
\texttt{<predicted\_level>Level-1</predicted\_level>}\\
The amount of money a certain company makes can be represented by the graph $y=x^2-8x-33$, where $x$ is the quantity produced. To find the smallest quantity at which the company can produce to break even or make a profit, we need to find the roots of the equation $x^2-8x-33=0$.

The quadratic equation $x^2-8x-33=0$ can be factored as $(x-11)(x+3)=0$. Therefore, the roots are $x=11$ and $x=-3$.

Since $x$ represents the quantity produced, it must be a non-negative value. Thus, the smallest quantity at which the company can produce to break even or make a profit is $x=11$.

The final answer is $\boxed{11}$.

\textbf{Solution to Problem 2:}\\
\texttt{<predicted\_level>Level-0</predicted\_level>}\\
If $f(x)=x+2$ and $g(x)=x^2$, then for what value of $x$ does $f(g(x))=g(f(x))$? Express your answer as a common fraction.

First, we find $f(g(x))$:
\[
f(g(x)) = f(x^2) = x^2 + 2.
\]
Next, we find $g(f(x))$:
\[
g(f(x)) = g(x+2) = (x+2)^2 = x^2 + 4x + 4.
\]
We set $f(g(x))$ equal to $g(f(x))$:
\[
x^2+2 = x^2+4x+4.
\]
Subtract $x^2$ from both sides:
\[
2 = 4x+4.
\]
Subtract 4 from both sides:
\[
-2 = 4x.
\]
Divide by 4:
\[
x=-\frac{1}{2}.
\]
The final answer is $\boxed{-\frac{1}{2}}$.

\textbf{Solution to Problem 3.}\\
\textit{The generation stops here before producing a predicted level or a boxed answer for Problem 3.}
\end{tcolorbox}

\begin{tcolorbox}[
    breakable,
    colback=green!5,
    colframe=green!40!black,
    title=\textbf{ROI-Reasoning Output},
    fonttitle=\bfseries,
    arc=1mm
]
\footnotesize
\textbf{Solution to Problem 1:}\\
\texttt{<predicted\_level>Level-0</predicted\_level>}\\
To find the break-even or make-a-profit point, we need to find the $x$-values where the profit is zero or positive. The profit is given by the equation $y=x^2-8x-33$. Setting $y=0$, we solve the quadratic equation $x^2-8x-33=0$. Factoring the quadratic equation, we get $(x-11)(x+3)=0$. The solutions are $x=11$ and $x=-3$. Since we are looking for the smallest quantity at which the company can produce to break even or make a profit, the answer is $\boxed{11}$.

\textbf{Solution to Problem 2:}\\
\texttt{<predicted\_level>Level-0</predicted\_level>}\\
We need to find the value of $x$ such that $f(g(x))=g(f(x))$. First, compute $f(g(x))$:
\[
f(g(x))=f(x^2)=x^2+2.
\]
Next, compute $g(f(x))$:
\[
g(f(x))=g(x+2)=(x+2)^2=x^2+4x+4.
\]
Setting them equal gives:
\[
x^2+2=x^2+4x+4.
\]
Simplifying, we get:
\[
2=4x+4 \implies 4x=-2 \implies x=-\frac{1}{2}.
\]
The answer is $\boxed{-\frac{1}{2}}$.

\textbf{Solution to Problem 3:}\\
\texttt{<predicted\_level>Level-0</predicted\_level>}\\
Let the number of boys be $2 \times \text{number of girls}$. Since there are 60 girls, the number of boys is $2 \times 60=120$. The total number of students is $120+60=180$. Since there are 5 students to every teacher, the number of teachers is $\frac{180}{5}=36$. The answer is $\boxed{36}$.
\end{tcolorbox}

\begin{tcolorbox}[
    colback=gray!5,
    colframe=gray!50!black,
    title=\textbf{Ground Truth},
    fonttitle=\bfseries,
    arc=1mm
]
\small
\textbf{Problem 1:} $\boxed{11}$ \quad
\textbf{Problem 2:} $\boxed{-\frac{1}{2}}$ \quad
\textbf{Problem 3:} $\boxed{36}$
\end{tcolorbox}

\paragraph{Observation.}
This example isolates the effect of adaptive reasoning length. The two models obtain the same correctness on the first two problems, so the gain is not caused by sacrificing early accuracy. Instead, RARL changes the reasoning trajectory within attempted problems: it lowers the first predicted level from Level-1 to Level-0 and reduces the first-two-problem output from 297 to 183 estimated tokens. The saved prefix budget is then used to complete Problem 3 correctly, whereas the MFT-only output is truncated before producing a complete boxed answer. Notably, the gain here does not come from an early Level-3 skip; it comes from shorter, still-correct reasoning on attempted problems. This example therefore illustrates that RARL improves the full reasoning trajectory under a global budget, rather than merely learning a binary solve-or-skip rule.

\subsection{Failure from Cost Underestimation}
\label{app:cost_underestimation_case}

To complement the successful case above, we include an illustrative Medium-512 failure case of ROI-Reasoning. Although the model achieves high aggregate predicted-level accuracy, local underestimation errors still occur. In this example, the model correctly skips the first AIME problem with \texttt{Level-3}, but predicts the second problem as \texttt{Level-0}. The resulting response expands beyond the expected short-solution range and is truncated before producing a boxed answer, so the model never reaches the third problem.

\paragraph{Execution summary.}
\begin{itemize}
    \item \textbf{ROI-Reasoning execution (failure case):}
    \begin{itemize}
        \item Q1: Level-3, skipped with $\boxed{NA}$.
        \item Q2: Level-0, but the response grows into a longer segment and is truncated before a boxed answer.
        \item Q3: never reached under the global budget.
    \end{itemize}
\end{itemize}

\begin{tcolorbox}[
    breakable,
    colback=gray!5,
    colframe=gray!50!black,
    title=\textbf{Input Problems (Medium-512)},
    fonttitle=\bfseries,
    arc=1mm
]
\small
\textbf{Problem 1.}
Let
\[
f(x)=\frac{(x-18)(x-72)(x-98)(x-k)}{x}.
\]
There exist exactly three positive real values of $k$ such that $f$ has a minimum at exactly two real values of $x$. Find the sum of these three values of $k$.

\textbf{Problem 2.}
A palindrome is a number that is the same when read forwards and backwards, such as $43234$. What is the smallest five-digit palindrome that is divisible by $11$?

\textbf{Problem 3.}
Joey plays football every week. Last week he played 2 matches on Monday, 1 match on Friday, and on Saturday he played double the number of matches he played on Monday. How many matches did Joey play in one week?
\end{tcolorbox}

\begin{table*}[h]
\centering
\setlength{\tabcolsep}{3pt}
\renewcommand{\arraystretch}{1}
\fontsize{7pt}{8.9pt}\selectfont

\resizebox{\textwidth}{!}{%
\begin{tabular}{lcccccccccccc}
\toprule
\multirow{3}[2]{*}{\textbf{Method}}
& \multicolumn{6}{c}{\textbf{Medium}}
& \multicolumn{6}{c}{\textbf{Hard}} \\
\cmidrule(lr){2-7} \cmidrule(lr){8-13}
& \multicolumn{3}{c}{1024}
& \multicolumn{3}{c}{1536}
& \multicolumn{3}{c}{1024}
& \multicolumn{3}{c}{1536} \\
\cmidrule(lr){2-4} \cmidrule(lr){5-7}
\cmidrule(lr){8-10} \cmidrule(lr){11-13}
& \textbf{Score} {\scriptsize$\uparrow$}
& Score$_{\mathrm{easy}}$
& \textbf{Regret} {\scriptsize$\downarrow$}
& \textbf{Score} {\scriptsize$\uparrow$}
& Score$_{\mathrm{easy}}$
& \textbf{Regret} {\scriptsize$\downarrow$}
& \textbf{Score} {\scriptsize$\uparrow$}
& Score$_{\mathrm{easy}}$
& \textbf{Regret} {\scriptsize$\downarrow$}
& \textbf{Score} {\scriptsize$\uparrow$}
& Score$_{\mathrm{easy}}$
& \textbf{Regret} {\scriptsize$\downarrow$} \\
\midrule

\multicolumn{13}{c}{\textbf{Open-Source Models (1.5B--7B)}} \\

\textbf{Qwen2.5-Math-7B-Instruct}
& 30.23 & 54.39 & 0.44
& \underline{45.81} & 61.26 & 0.25
& 20.43 & 50.80 & 0.60
& \underline{38.91} & 58.07 & 0.33 \\

\textbf{Qwen2.5-7B-Instruct}
& \textbf{40.66} & 55.12 & 0.26
& \textbf{53.02} & 57.42 & 0.08
& \underline{32.65} & 51.65 & 0.37
& \textbf{48.55} & 55.07 & 0.12 \\

\textbf{Qwen2.5-Math-1.5B-Instruct}
& 29.04 & 47.90 & 0.39
& 39.18 & 52.55 & 0.25
& 21.38 & 45.08 & 0.53
& 32.27 & 49.95 & 0.35 \\

\textbf{Qwen2.5-1.5B-Instruct}
& 18.66 & 24.01 & 0.22
& 22.72 & 23.81 & \underline{0.05}
& 14.87 & 22.77 & 0.35
& 19.91 & 23.00 & 0.13 \\

\midrule
\multicolumn{13}{c}{\textbf{Meta-Cognitive Methods}} \\

\textbf{Qwen2.5-1.5B-Instruct}
& & & & & &
& & & & & & \\

\quad + Plan-and-Solve
& 14.04 & 22.45 & 0.37
& 18.02 & 25.81 & 0.30
& 9.39 & 20.34 & 0.54
& 14.54 & 23.08 & 0.37 \\

\quad + MFT
& 32.91 & 35.26 & \underline{0.07}
& 33.86 & 35.95 & 0.06
& 30.37 & 33.26 & \underline{0.09}
& 31.36 & 33.95 & \underline{0.08} \\

\rowcolor[RGB]{236,244,252}
\quad \textbf{+ ROI-Reasoning}
& \underline{39.49} & 40.65 & \textbf{0.03}
& 39.85 & 40.60 & \textbf{0.02}
& \textbf{37.23} & 37.90 & \textbf{0.02}
& 37.50 & 38.51 & \textbf{0.03} \\

\bottomrule
\end{tabular}}
\caption{
Unseen-$N$ generalization for $N=4$ after training with $N=3$.
Scores are normalized to a 0--100 scale. We report Score,
Score$_{\mathrm{easy}}$, and Reference-Order Regret on Medium and
Hard papers under 1024- and 1536-token budgets. Bold and underlined
values denote the best and second-best Score and Regret, respectively.
}
\label{tab:unseen_n4}
\end{table*}

\begin{table*}[h]
\centering
\setlength{\tabcolsep}{3pt}
\renewcommand{\arraystretch}{1}
\fontsize{7pt}{8.9pt}\selectfont

\resizebox{\textwidth}{!}{%
\begin{tabular}{lcccccccccccc}
\toprule
\multirow{3}[2]{*}{\textbf{Method}}
& \multicolumn{6}{c}{\textbf{Medium}}
& \multicolumn{6}{c}{\textbf{Hard}} \\
\cmidrule(lr){2-7} \cmidrule(lr){8-13}
& \multicolumn{3}{c}{1024}
& \multicolumn{3}{c}{1536}
& \multicolumn{3}{c}{1024}
& \multicolumn{3}{c}{1536} \\
\cmidrule(lr){2-4} \cmidrule(lr){5-7}
\cmidrule(lr){8-10} \cmidrule(lr){11-13}
& \textbf{Score} {\scriptsize$\uparrow$}
& Score$_{\mathrm{easy}}$
& \textbf{Regret} {\scriptsize$\downarrow$}
& \textbf{Score} {\scriptsize$\uparrow$}
& Score$_{\mathrm{easy}}$
& \textbf{Regret} {\scriptsize$\downarrow$}
& \textbf{Score} {\scriptsize$\uparrow$}
& Score$_{\mathrm{easy}}$
& \textbf{Regret} {\scriptsize$\downarrow$}
& \textbf{Score} {\scriptsize$\uparrow$}
& Score$_{\mathrm{easy}}$
& \textbf{Regret} {\scriptsize$\downarrow$} \\
\midrule

\multicolumn{13}{c}{\textbf{Open-Source Models (1.5B--7B)}} \\

\textbf{Qwen2.5-Math-7B-Instruct}
& 26.01 & 51.97 & 0.50
& \underline{41.28} & 61.44 & 0.33
& 16.52 & 44.12 & 0.63
& 32.02 & 54.94 & 0.42 \\

\textbf{Qwen2.5-7B-Instruct}
& \underline{37.85} & 55.37 & 0.32
& \textbf{52.79} & 59.19 & 0.11
& \underline{28.77} & 48.74 & 0.41
& \textbf{46.47} & 54.40 & 0.15 \\

\textbf{Qwen2.5-Math-1.5B-Instruct}
& 23.91 & 44.08 & 0.46
& 33.94 & 50.62 & 0.33
& 17.12 & 38.55 & 0.56
& 26.70 & 46.42 & 0.42 \\

\textbf{Qwen2.5-1.5B-Instruct}
& 16.37 & 22.69 & 0.28
& 20.66 & 23.50 & 0.12
& 12.64 & 20.78 & 0.39
& 17.80 & 21.98 & 0.19 \\

\midrule
\multicolumn{13}{c}{\textbf{Meta-Cognitive Methods}} \\

\textbf{Qwen2.5-1.5B-Instruct}
& & & & & &
& & & & & & \\

\quad + Plan-and-Solve
& 12.56 & 21.52 & 0.42
& 19.00 & 26.48 & 0.28
& 7.82 & 17.65 & 0.56
& 14.08 & 22.73 & 0.38 \\

\quad + MFT
& 31.71 & 34.77 & \underline{0.09}
& 33.29 & 35.27 & \underline{0.06}
& 27.82 & 30.28 & \underline{0.08}
& 29.10 & 31.33 & \underline{0.07} \\

\rowcolor[RGB]{236,244,252}
\quad \textbf{+ ROI-Reasoning}
& \textbf{40.56} & 41.86 & \textbf{0.03}
& 41.04 & 41.95 & \textbf{0.02}
& \textbf{35.39} & 36.54 & \textbf{0.03}
& \underline{36.16} & 37.07 & \textbf{0.02} \\

\bottomrule
\end{tabular}}
\caption{
Unseen-$N$ generalization for $N=5$ after training with $N=3$.
Scores are normalized to a 0--100 scale. We report Score,
Score$_{\mathrm{easy}}$, and Reference-Order Regret on Medium and
Hard papers under 1024- and 1536-token budgets. Bold and underlined
values denote the best and second-best Score and Regret, respectively.
}
\label{tab:unseen_n5}
\end{table*}
            
\paragraph{Observation.}
This case shows that improved global planning does not eliminate all local cost-prediction errors. The model makes a reasonable high-level decision on Problem 1, but underestimates the cost of Problem 2 and spends too much of the remaining budget before reaching a boxed answer. As a result, it never reaches the final GSM8K problem. This failure case complements the successful case above by showing that residual regret can still arise from cost underestimation within an attempted problem.

\begin{tcolorbox}[
    breakable,
    colback=red!5,
    colframe=red!45!black,
    title=\textbf{ROI-Reasoning Output},
    fonttitle=\bfseries,
    arc=1mm
]
\footnotesize
\textbf{Solution to Problem 1:}\\
\texttt{<predicted\_level>Level-3</predicted\_level>}\\
$\boxed{NA}$

\textbf{Solution to Problem 2:}\\
\texttt{<predicted\_level>Level-0</predicted\_level>}\\
A five-digit palindrome has the form $abcba$, where $a, b, c$ are digits and $a \neq 0$. To be divisible by 11, the alternating sum of the digits must be divisible by 11. That is, $a-b+c$ must be divisible by 11. The smallest five-digit palindrome is 10001. We check the alternating sum: $1-0+0-0+1=1$, which is not divisible by 11. The next smallest is 11111, and the alternating sum is $1-1+1-1+1=1$, which is not divisible by 11. The next is 12221, and the alternating sum is $1-2+2-2+1=-2$, which is not divisible by 11. The next is 13331, and the alternating sum is $1-3+3-3+1=-1$, which is not divisible by 11. The next is 14441, and the alternating sum is $1-4+4-4+1=-2$, which is not divisible by 11. The next is 15551, and the alternating sum is $1-5+5-5+1=-2$, which is not divisible by 11. The next is 16661, and the alternating sum is $1-6+6-6+1=-1$, which is not divisible by 11. The next is 17771, and the alternating sum is $1-7+7-7+1=-2$, which is not divisible by 11. The next is 18881, and the alternating sum is $1-8+8-8+1=-1$, which is not divisible by 11. The next is
\end{tcolorbox}

The generation stops here before producing a boxed answer for Problem 2 or any answer for Problem 3.

\begin{tcolorbox}[      
    colback=gray!5,
    colframe=gray!50!black,
    title=\textbf{Ground Truth},
    fonttitle=\bfseries,
    arc=1mm
]
\small
\textbf{Problem 1:} $\boxed{240}$ \quad
\textbf{Problem 2:} $\boxed{10901}$ \quad
\textbf{Problem 3:} $\boxed{7}$
\end{tcolorbox}

\section{Generalization to Unseen Sequence Lengths}
\label{app:unseen_sequence_lengths}

The main model is trained only on three-problem papers with a
1024-token budget. Tables~\ref{tab:unseen_n4}
and~\ref{tab:unseen_n5} evaluate whether the learned allocation
behavior transfers to four- and five-problem papers under both the
training budget and an unseen 1536-token budget. To enable comparison
across sequence lengths, scores are divided by the maximum attainable
score for the corresponding paper length and reported on a 0--100
scale.

\paragraph{Unseen-\texorpdfstring{$N$}{N} Paper Construction.}
Because the available pool of easy problems is limited, the Medium
and Hard papers in the unseen-$N$ evaluation are constructed
independently rather than by reordering the same underlying problem
sets. For each setting, Score$_{\mathrm{easy}}$ is computed by
reordering that setting's own problems from easy to hard, and
reference-order regret is calculated against this setting-specific
reference. Therefore, Medium and Hard do not share
Score$_{\mathrm{easy}}$ or regret. All models are evaluated on the
same papers within each setting, so method comparisons remain
controlled.

\paragraph{Results.} For $N=4$, ROI-Reasoning improves over the same-backbone MFT model in every
paper type and budget. The gain persists under the larger 1536-token
budget, showing that improved tight-budget allocation does not
sacrifice performance when more computation is available.
Reference-order regret remains at or below 0.03.

For $N=5$, ROI-Reasoning again improves over MFT across all four settings
and keeps reference-order regret between 0.02 and 0.03. Larger 7B
baselines retain higher absolute scores in some Medium-1536 settings,
so these results support robust same-backbone generalization rather
than universal superiority over larger models.

\end{document}